%% file: _IsoClustering-Arxiv-91-1-1.tex
\documentclass[journal]{IEEEtran}
\usepackage{cite}
\usepackage{calc}
\usepackage{algorithm}
\usepackage{algorithmic}
\usepackage{graphicx}
\graphicspath{{/images}}
\usepackage{amssymb,mathrsfs}
\usepackage[cmex10]{amsmath}
\usepackage[tight,footnotesize]{subfigure}
\usepackage{url}
\input{environments}

\begin{document}
\title{Clustering Using  Isoperimetric Number of Trees}

\author{Amir~Daneshgar, %~\IEEEmembership{Member,~IEEE,}
        Ramin~Javadi %~\IEEEmembership{Fellow,~OSA,}
        and~Basir~Shariat~Razavi%~\IEEEmembership{Life~Fellow,~IEEE}% <-this % stops a space}

\thanks{A. Daneshgar is with the Department
of Mathematical Sciences, Sharif University of Technology, Tehran,
 P.O. Box {\rm 11155--9415} Iran. e-mail: (\href{mailto:daneshgar@sharif.ac.ir}{\tt daneshgar@sharif.ir}).}
 % <-this % stops a space

\thanks{R. Javadi is with the Department
of Mathematical Sciences, Isfahan University of Technology, Isfahan,
 P.O. Box {\rm 8415683111} Iran. e-mail: (\href{mailto:rjavadi@cc.iut.ac.ir}{\tt rjavadi@cc.iut.ac.ir}).}
 % <-this % stops a space
\thanks{B. Shariat Razavi is with the Department
of Mathematical Sciences, Sharif University of Technology, Tehran,
 P.O. Box {\rm 11155--9415} Iran. e-mail: (\href{mailto: basirshariat@alum.sharif.ir}{\tt  basirshariat@alum.sharif.ir}).}
 % <-this % stops a space
%\\

%\thanks{Manuscript received ???; revised ???.}
}

% The paper headers
%\markboth{IEEE TRANSACTIONS ON PATTERN ANALYSIS AND MACHINE INTELLIGENCE,,~Vol.~??, No.~??, ???}%
%{Shell \MakeLowercase{\textit{et al.}}: Bare Demo of IEEEtran.cls for Journals}
% make the title area
\maketitle

\begin{abstract}

In this paper we propose a graph-based data clustering algorithm which is based on exact clustering of a minimum spanning tree in terms of a minimum isoperimetry criteria. We show that our basic clustering algorithm  runs in 
$O(n \log n)$ and with post-processing in $O(n^2)$ (worst case) time where $n$ is the size of the data set. We also show that our generalized graph model 
which also allows the use of potentials at vertices can be used to extract a more detailed pack of information as the {\it outlier profile} of the data set. In this direction we show that our approach can be used to define the concept of an outlier-set in a precise way and we propose approximation algorithms for finding such sets.
We also provide a comparative performance analysis of our algorithm with other related ones and we show that the  new clustering algorithm (without the outlier extraction procedure) behaves quite effectively even on hard 
benchmarks and handmade examples.

\end{abstract}
\IEEEpeerreviewmaketitle

\input{relatedworks}
\input{segmentationalg}
\input{performanceanalysis}

\input{outlierprofile}

\appendix[Proof of Proposition~\ref{pro:residue}]
\input{appendix}
%\section*{Acknowledgment}

%The authors would like to thank...

\ifCLASSOPTIONcaptionsoff
  \newpage
\fi
%\bibliography{bibliography}
%\bibliographystyle{IEEEtran}

% Generated by IEEEtran.bst, version: 1.13 (2008/09/30)

\end{document}

%% file: environments.tex
\newcounter{theorem}
\newcommand{\environ}[2]{\vspace{1pt}\refstepcounter{theorem}\par\noindent  \textbf{#2 \thetheorem #1.} \\[2pt] }
\newenvironment{thm}[1][]{\environ{#1}{Theorem} \it}{\par}

\newenvironment{prop}[1][]{\environ{#1}{Proposition} \it}{\par}
\newenvironment{defin}[1][]{\environ{#1}{Definition} \rm}{\par}
\newenvironment{prob}[3]{\vspace{2pt}
\par\noindent {\bf #1}\\[5pt]
\hspace*{-6pt}\begin{tabular}{ll}
{INSTANCE:}& \parbox[t]{\linewidth-2.3cm}{#2}\\[10pt]
 QUERY:    & \parbox[t]{\linewidth-2.3cm}{#3}}
{\end{tabular}\\[5pt]}

\providecommand{\MR}{\relax\ifhmode\unskip\space\fi MR }

\providecommand{\href}[2]{#2}

%% file: relatedworks.tex
\section{Introduction}

\subsection{A concise survey of main results}
{\Large D}ata clustering, as the unsupervised grouping of similar patterns into clusters, is a central problem in engineering disciplines and applied sciences which is also constantly under theoretical and practical  development and verification. In this article we are concerned with graph based data clustering  methods which are extensively studied and developed mainly because of their simple implementation and acceptable efficiency in a number of different fields as signal and image processing, computer vision,
computational biology, machine learning and networking to name a few.

The main contribution in this article can be described as a general graph-based data clustering algorithm which falls into the category of such algorithms that use a properly defined sparsest cut problem as the clustering criteria. In this regard, it is instructive to note some highlights of our approach before we delve into the details in subsequent sections (details of our approach as well as a survey of related contributions will appear in the second part of this introduction).

 It has been already verified that graph-based clustering methods that operate in terms of non-normalized cuts are not suitable for general data clustering and behave poorly in comparison to the normalized versions (e.g. see \cite{SHIMAL00}). Moreover, it is well known that there is a close relationship between the minimizers of the normalized cut problem, spectral clustering solutions, mixing rates of random walks, the minimizers of the K-means cost function, kernel PCA and low dimensional embedding, while the corresponding decision problems are known to be NP-complete in general (e.g. see \cite{DAM10,JCTB,ARRAVA08,LITI07,NGJOWE01,BEDEROPA04} and references therein).

 In this article, we will provide an efficient clustering algorithm  which is based on a relaxation of the feasible space of solutions from the set of {\it partitions} to the larger set of {\it subpartitions} (i.e. mutually disjoint subsets of the domain).
 From one point of view, our algorithm can be considered as a generalization of Grady and Schwartz approach \cite{GRSCH06,GRADY06} based on isoperimetry problems while we extensively rely on the results of \cite{DAM10} and \cite{JCTB}. Also, we believe that this relaxation which is based on moving from the space of partitions to the space of subpartitions not only provides a chance of making the problem easier to solve but also is in coherence with the natural phenomena of having undesirable data or outliers. We will use this property to show that our algorithm can be enhanced to a more advanced procedure which is capable of presenting a hierarchy of data similarity profile which can lead to the extraction of outliers.
 
 In this regard, one may comment on some different aspects of this approach as follows.

{\bf Theoretical aspects}: From a theoretical point of view, it is proved in \cite{JCTB} that the normalized cut criteria is not formally well defined in the sense that it does not admit a variational description through a real function relaxation of the problem (i.e. it does not admit a Federer-Fleming type theorem). However, for $k \geq 2$, the well-defined version, known as  the {\it $k$-isoperimetry problem} (defined in \cite{JCTB}), whose  definition is in terms of normalized-flow minimization on $k$-subpartitions, actually admits such a relaxation. It should be noted that although there are some approaches to clustering which are based on the classical $2$-isoperimetry (i.e. Cheeger constant) on weighted graphs (e.g. see \cite{GRSCH06}), but as it follows from the results of \cite{JCTB}, in the classical model the difference between the cases of partitions and subpartitions only is observable when $k \geq 3$, and consequently, our approach is  completely different in nature from iterative $2$-partitioning or spectral approximation methods based on eigenmaps already existing in the literature.

    Also, as a bit of a surprise (see Theorem~\ref{thm:effcomputisotree}), it turns out that a special version of the $k$-isoperimetry problem is {\it efficiently solvable} for trees.   This fact along with a well-known approach of finding an approximate graph partitioning through minimum spanning trees constitute the core of our algorithm.
    
{\bf Practical aspects}:
There are different practical aspects of the proposed algorithm that one may comment on. Firstly, the proposed approximation algorithm run-time is almost linear in terms of the size of input-data which provides an opportunity to cluster large data sets. Also, it should be noted that our
algorithm for $k$-clustering obtains an exact optimal clustering of a suitably chosen subtree in a global approach and does not apply an iterative two-partitioning or an approximation through eigenmaps.
This in a way is one of the reasons supporting a better approximation of our algorithm compared to the other existing ones. In this regard, we also present a number of experimental results justifying a better performance of our algorithm in practice (see Tables~\ref{tab:uci}, \ref{tab:uci2} and Section~\ref{sec:tests}).

Secondly, we should note that approximation through the isoperimetry criteria provides an extra piece of information  as a (possibly nonempty) subset of the domain (since the union of subpartitions may not be a covering). This piece of information  makes it possible to obtain the almost minimal clustering as well as to extract deviated data and outliers, at the same time. In order to handle this extra information, we have generalized our graph model to the case of a {\it weighted graph with potential}.
This generalization of the graph representation model is another original aspect of our contribution where
we rely on results of \cite{DAM10} and \cite{JCTB} in this {\it more general setting} (see Theorem~\ref{thm:effcomputisotree}). It is interesting to note that in this more general setting our results presented in Section~\ref{sec:outlierprofile} show that not only we can handle the case of outlier extraction with clustering at the same time, but also the new set up will make it possible to extract outliers even in the case of $2$-clusterings (which is theoretically meaningless by definition when one is using the classical isoperimetry or $2$-normalized cuts). Using this setting we propose a formal definition for the {\it outlier profile} of a data set and,  moreover, we provide a couple of examples to study the efficiency of the proposed method in extracting outliers.

\subsection{Background and related contributions}
Unsupervised grouping of data based on a predefined similarity criteria is usually referred to as {\it data clustering} in general, where in some more specific applications one may encounter some other terms as {\it segmentation} in image processing or {\it grouping} in data mining.
Based on its importance and applicability, there exists a very vast literature related to this subject (e.g. see \cite{JAMUFL99,DEEMFI06} for some general background), however, in this article we are mainly concerned with clustering algorithms that rely on a representation of data as a simple weighted graph in which the edge-weights are tuned, using a predefined similarity measure (e.g. see Section~\ref{sec:themodel}, \cite{SCH07} and references therein).

Graph-based data clustering is usually reduced to the {\it graph partitioning} problem on the corresponding weighted graph which is also well-studied in the literature. To this end, it is instructive to note that from this point of view and if one considers a weighted graph as a geometric object, then the partitioning problem can be linked to a couple of very central and extensively studied problems in geometry as {\it isoperimetry problem}, {\it concentration of measure} and {\it estimation of diffusion rates} (e.g. see \cite{JCTB,ARRAVA08} and references therein).

A graph-based clustering or a graph partitioning problem is usually reduced to an optimization problem where the cost function is a measure of sparsity or density related to the corresponding classes of data. From this point of view, it is not a surprise to see a variety of such measures in the literature, however, from a more theoretical standpoint such similarity measures are well-studied and, at least, the most geometrically-important classes of them are characterized (e.g. see \cite{Rot85} for a very general setting). In this context such measures usually appear as {\it norms} or their normalized versions that should be minimized or maximized to lead to the expected answer.

What is commonly refereed to as {\it spectral clustering} is the case in which the corresponding normalized norm is expressed as an $L^2$ (i.e. Euclidean) norm and admits a real-function relaxation whose minimum is actually an eigenvalue of the weight (or a related) matrix of the graph. This special case along with the important fact that, the spectral properties (i.e. eigenvalues and eigenfunctions) of a finite matrix can be effectively (at most in $O(n^3)$ time) computed, provides a very interesting setting for data clustering in which the corresponding optimization problem can be tackled with
using the well-known tools of linear algebra and operator theory (e.g. see \cite{KARAVEVE04,LUBEBO08,LUX07,WEI99,VEM07,KAVE08,LRTV11,YUSH03} and references therein for a general background in spectral methods).

Although, applying spectral methods are quite effective and vastly applied in data clustering, but still the time complexity of the known algorithms
and also the approximation factor of this approach in not as good as one expects when one is dealing with large data sets (e.g. see \cite{PEWE07} that proves an approximation factor of at most $2$). On the other way round, these facts leads one to consider the original normalized versions of the $L^1$ norm that reduces clustering to the sparsest (or similar minimal) cut problems or their real-function relaxations as the corresponding approximations. It is proved in \cite{JCTB} that the most natural such normalized norms {\it do not} admit real-function relaxations when they are minimized over partitions of their domain. Moreover, it is shown in the same reference that such normalized norms {\it do admit} such real-function relaxations when they are minimized over subpartitions of their domain. In this new setting the minimum values, that correspond to the eigenvalues in the spectral $L^2$ setting, are usually referred to as {\it isoperimetric constants}.

Unfortunately, contrary to the case of $L^2$, decision problems corresponding to the isoperimetry problems are usually NP-hard (e.g. see \cite{SHIMAL00,DAM10,LITI07,J85}), which shows that computing the exact value of the isoperimetric constants is not an easy task. There has been a number of contributions in the literature whose main objectives can be described as to proposing different methods to get around this hardness problem and find an approximation for the corresponding isoperimetry problem as a criteria of clustering, and consequently, obtaining an approximate clustering of the given data.

In this regard, one may at least note two different approaches as follows. In the one hand, there has been  contributions who has tried to reduce the problem to the more tractable case of trees by first finding a suitable subtree of the graph and then try to approximately cluster the tree itself (e.g. see \cite{Asano:1988:CAB:73393.73419,Paivinen2005921,Grygorash:2006:MST:1190614.1191118,10.1109/TKDE.2005.112,flake04graphclust,Xu199747,WA08}). The difference between such contributions usually falls into the way of choosing the subtree and the method of their clustering. On the other hand,
one may also try to obtain a global clustering by a mimic of spectral methods through solving not an eigenfunction problem but a similar problem in $L^1$ (e.g. see \cite{GRSCH06,GRADY06}). These methods usually follow an iterative $2$-partitioning since there was not much information about approximations
for higher order eigenfunctions or similar solutions in $L^1$ until recently (e.g. see \cite{LITI07,GRSCH06}).

Our main contribution in this article can be described as a culmination of above mentioned ideas that strongly rely on some recent studies of higher order solutions
of isoperimetry problems (see \cite{DAM10,JCTB}), in which we first search for a suitable spanning subtree and after that we obtain the {\it exact} solution of the corresponding optimization problem for our suitably chosen isoperimetric constant (see Section~\ref{sec:themodel}). Also, in this setting we will obtain a
subset of unused data given as the complement of the obtained clustering as a subpartition and we will try to analyze this {\it extra} output of our algorithm as an outlier detection procedure (see Section~\ref{sec:outlierprofile}). To do this we adopt a generalize  graph model as a {\it weighted graph with potential} and we will
be needing a generalization of some results of \cite{DAM10} and \cite{JCTB} that will be presented in Section~\ref{sec:themodel}. Also, in Sections~\ref{sec:tests} and \ref{sec:outlierprofile} we provide  experimental results to show the efficiency and the performance of our proposed algorithms.

%% file: segmentationalg.tex
\newcommand{\w}{\omega}
\newcommand{\ph}{\varphi}
\newcommand{\p}{p}
\newcommand{\sP}{\mathscr{P}}
\newcommand{\sD}{\mathscr{D}}
\newcommand{\cF}{\mathscr{F}}
\newcommand{\cO}{\mathscr{O}}
\newcommand{\cA}{\mathcal{A}}
\newcommand{\cB}{\mathcal{B}}
\newcommand{\mnc}{{\rm MNC}}
\newcommand{\iso}{{\rm MISO}}
\newcommand{\pa}{\partial}
\section{The clustering model and algorithm}\label{sec:themodel}
In this section we introduce our graph based model and the proposed  clustering algorithm.

Fix positive integers $d \geq 1$, $k$ and $n$ such that  $2\leq k\leq n$ and
let $X=\{x_1,\ldots,x_n\}$ be a set of $n$ vectors in $\mathbb{R}^d$.
A standard $k$-clustering problem for $X$ is to find a $k$-partition of $X$ with a high intra-clusters similarity as well as a low inter-clusters similarity (with respect to a predefined similarity measure).

In graph-based methods of clustering, the data-set $X$ is represented by a weighted graph on $n$ vertices, where each vertex corresponds to a vector in $X$ and the weight of an edge $x_{i}x_{j}$ reflects the similarity between vectors $x_{i}$ and $x_{j}$. In this article, our graph model consists of a simple graph $G=(X,E)$ on the vertex set $X:=\{x_1,\ldots, x_n\}$ endowed with three weight functions, namely, a vertex-weight function $\w:X\to \mathbb{R}^+$, an edge-weight function $\ph: E\to \mathbb{R}^+$ called the {\it flow} and a function $\p:X\to \mathbb{R}$ called the {\it potential}. The function $\w$ is used for the weight of each element of the data-set $X$ and the similarities between pairs of elements are denoted by the flow function $\ph$. The potential of a vertex $x_{i} \in X$, $p(x_i)$, is used to represent the extent of {\it isolation} or {\it alienation} of $x_i$ from other elements. In this setting, the weighted graph  is called the {\it similarity} or {\it affinity} graph and is denoted by $(G,\w,\ph,\p)$ where, hereafter, the size of the data-set which is equal to the number of vertices is fixed to be $n:=|X|$. For instance, in a classic way of modelling similarities, one may define the flow and vertex weight functions as follows,
\begin{equation}
\begin{split}
\label{eq:simgraph}
\forall\  1\leq i\neq j\leq n, \ \ph(x_ix_j)&:= \exp(-\| x_i-x_j\|_{_{2}}/ 2\sigma^2),\\
\forall\ \leq i\leq n,\ \w(x_i)&:= \sum_{j=1}^n \ph(x_ix_j).
\end{split}
\end{equation}

We do not elaborate on the more or less complex subject of proposing suitable methods for graphical presentation of data sets. The interested reader is referred to the existing literature (e.g. \cite{ZEPE04}) to see how the scale parameter 
$\sigma$ is chosen and how it affects the performance of graph based algorithms. However, in order to present a complete comparison in Section~\ref{sec:tests} we consider both {\it global scaling} and {\it local scaling}
models in our performance analysis. 

In the sequel we will discuss the important role of potentials in Section~\ref{sec:outlierprofile}, when we elaborate on the capability of our algorithm to detect the outlier profile of the data set.

To describe our model of clustering, first we need to define a couple of notations. For a subset $A\subset X$, the {\it boundary} of $A$, denoted by $\pa A$, is defined as,
\[\pa A:=\{e=xy\in E\ |\ x\in A, y\in X\setminus A\}.\]
Also, for any given finite set $A$, a function $f:A\to \mathbb{R}$ and a subset $B \subseteq A$, we define
\[f(B):=\sum_{x\in B} f(x).\]
The collection of all $k$-partitions of the set $X$ is denoted by $\sP_k(X)$. Given a weighted graph $(G,\w,\ph,\p)$ and an integer $2\leq k\leq n$, (the maximum version of) \textit{the $k$-normalized cut problem} seeks for a $k$-partition $\cA:=\{A_1,\ldots,A_k\}\in \sP_k(X)$ that minimizes the following cost function,
\begin{equation}\label{eq:cost}
{\rm cost}(\cA):= \max_{1\leq i\leq k} \frac{\ph(\pa A_i)+\p(A_i)}{\w(A_i)}.
\end{equation}
We define
\[\mnc_k(G):= \min_{\cA\in \sP_k(X)} {\rm cost}(\cA).\]
The quotient ${(\ph(\pa A_i)+\p(A_i))}/{\w(A_i)}$ is called \textit{normalized flow} of the set $A_i$.
The (max) normalized cut problem is known to be $NP$-hard for general graphs even when $k=2$ \cite{Mohar}. In  \cite{DAM10}  the same problem is investigated for weighted trees and it is proved that the corresponding decision problem for arbitrary $k$ remains $NP$-complete even for simple (unweighted) trees. In the same reference  a tractable relaxation of the problem is also proposed which is based on the relaxation of the feasible set from the set of $k$-partitions to the set of {\it $k$-subpartitions}, i.e. $k$ disjoint subsets of the vertex set. The set of all $k$-subpartitions of $X$, denoted by $\sD_k(X)$, is defined as follows,
\begin{align*}
\sD_k(X):= &\{\cA:=\{A_1,\ldots, A_k\}\ |\ \\
&\forall\ i, A_i\subset X, {\rm and } \ \forall\ i\neq j, A_i\cap A_j=\emptyset\}.
\end{align*}
For a subpartition $\cA=\{A_1,\ldots, A_k\}\in \sD_k(X)$, each element in $X\setminus \cup_{i=1}^k A_i$ is called a \textit{residue element} (w.r.t. $\cA$) and the number of residue elements is called the \textit{residue number} of $\cA$.

The  (maximum version) of {\it isoperimetric problem} seeks for a minimizer of the cost function in (\ref{eq:cost}) over the space of all $k$-subpartitions of $X$. We denote the minimum by $\iso_k(G)$, i.e.
\[\iso_k(G):= \min_{\cA\in \sD_k(X)} {\rm cost}(\cA).\]
It is not hard to check that there exist instances where the minimum $\iso_k(G)$ occurs on a subpartition which is not actually a partition, and also it can be verified that $\mnc_2(G)=\iso_2(G)$ when the potential function is equal to zero (see \cite{JCTB}).

The idea of relaxing the normalized cut problem to the isoperimetric problem have a number of justifications from  different points of view. On the one hand, from a computational viewpoint, the isoperimetric problem is more tractable in some special cases (e.g. see Theorem~\ref{thm:effcomputisotree}), while, on the other hand, from  a theoretical viewpoint, it can be verified that the isoperimetry problem admits a Federer-Fleming-type theorem, while the normalized cut problem doesn't satisfy such a relaxation in general (see Theorem~\ref{thm:FFT} and \cite{JCTB}). 

For two given real functions $f$ and $g$ on $X$ and a positive weight fuction $\w: X \to \mathbb{R}^+$ define the weighted inner product as 
$$\langle f_i,f_j \rangle_{\w} := \displaystyle{\sum_{x \in X}} f(x)g(x)\w(x).$$
Also, if $\cF^+(X)$ is the set of all non-negative real functions on $X$ and $k > 1$ is a positive integer,  $\cO_{_{k}}^+(X)$ stands for the set of $k$ mutually orthogonal functions in $\cF^+(X)$, i.e.
\begin{align*}
\cO_{_{k}}^+(X)&:=\{\{f_1,\ldots,f_k\}\ |\ f_i\in \cF^+(X),\\
&\quad \quad \quad \langle f_i,f_j \rangle_\w=0, \ \forall \ i\neq j\}.
\end{align*}
Now, one may verify that the following Federer-Fleming-type theorem holds. We deliberately exclude the proof since it is essentially a straight forward generalization of the proof already presented in \cite{JCTB} for the standard case (i.e. when the potential function is equal to zero).
\begin{thm}[{\cite{RAMINTHESIS}}]\label{thm:FFT}
For every weighted graph $(G,\w,\ph,p)$ and integer $k$,
\begin{equation*}
\begin{split}
\iso &_k(G)=
\displaystyle{\inf_{
\{f_{i}\}^{k}_{1} \in  {\cO}^+_{k}(G) }} \
\max_{1\leq i\leq k}\\[5pt]
&\left ( \frac{\displaystyle\sum_{xy\in E} \varphi(xy)\ |f_i(x)-f_i(y)|+ \sum_{x\in V} p(x)\ |f_i(x)|}{\displaystyle\sum_{x\in V} \w(x)\ |f_i(x)|} \right )
\end{split}
\end{equation*}
\end{thm}
It is shown in \cite{DAM10} that in the case of weighted trees, despite intractability of the normalized cut problem, the decision problem related to $\iso_k(G)$ is efficiently solvable in the following sense.
\begin{thm} [{\cite{DAM10}}]\label{thm:DAM}
For every weighted tree $(T,\w,\ph,\p)$, the decision version of the (max) isoperimetry problem can be efficiently solved in linear time.
\end{thm}
Since our proposed clustering method is based on the algorithm announced in Theorem~\ref{thm:DAM}, we include the algorithm  for completeness (see Algorithm~\ref{alg:daneshgar}). In what follows assume that a vertex $v$ is selected as the root and the vertices are ordered in a BFS order, as $x_1,\ldots,x_{n}=v$.
\begin{algorithm}[H]
\caption{Given a weighted tree $(T,\w,\ph,\p)$, an integer $k$ and a rational number $N$, decide whether there exists some $\cA\in \sD_k(X)$ such that ${\rm cost}(\cA)\leq N$ as in (\ref{eq:cost}).}
\label{alg:daneshgar}
 \begin{algorithmic}
 \STATE Initialize the set function $\eta: X\to \mathcal{P}(X)$ by $\eta(x_i):=\{x_i\}$ for each $1\leq i\leq n$.
 \STATE Define $i=j:=1$.
 \WHILE{$j<k$ \AND $i\leq n$}
 \STATE Let $u$ be the unique parent of $x_i$ and $e:=ux_i\in E$ (if $i=n$, then define $\ph(e):=0$)
 \IF{$\p(x_i)+\ph(e)\leq N \w(x_i)$}
 \STATE $j\leftarrow j+1$, $A_j\leftarrow\eta(x_i)$,  
 \STATE $\w(A_j)\leftarrow\w(x_i)$,
 \STATE $\ph(\pa A_j)\leftarrow \ph(e)+\p(x_i)$,
  \STATE $\p(u)\leftarrow\p(u)+\ph(e)$.
 \ELSIF{$\p(x_i)-\ph(e)<N \w(x_i)$}
 \STATE $\eta(u)\leftarrow \eta(u)\cup \eta(x_i)$,
 \STATE $\w(u)\leftarrow \w(u)+\w(x_i)$,
 \STATE $\p(u)\leftarrow\p(u)+\p(x_i)$.
 \ELSE[i.e. $\p(x_i)-\ph(e)\geq N \w(x_i)$]
 \STATE $\p(u)\leftarrow\p(u)+\ph(e)$
 \ENDIF
 \ENDWHILE
 \IF{$j=k$}
 \STATE \RETURN YES and $\{A_1,\ldots,A_k\}$
 \ELSE
 \STATE \RETURN NO
 \ENDIF
 \end{algorithmic}
\end{algorithm}

In this article we provide an improved version of Theorem~\ref{thm:DAM} as follows.
\begin{thm}\label{thm:effcomputisotree}
For every weighted tree $(T,\w,\ph,\p)$ on $n$ vertices and every integer $2\leq k\leq n$, the value of $\iso_k(T)$ and a minimizer in $\sD_k(V)$ can be found in time $O(n\log n)$.
\end{thm}
\begin{IEEEproof}
Let $T=(V,E,\w,\ph,\p)$ be a fixed weighted tree on $n$ vertices and $1\leq k\leq n$ be a fixed integer. Without loss of generality, assume that all the weights are integer. Define $$\w_*:=\min_{x\in V} \w(x),\quad \w^*:=\sum_{x\in V} \w(x),$$ 
$$\ph_*:=\min_{e\in E} \ph(e),\quad \ph^*:=\sum_{e\in E} \ph(e),$$ 
$$\p_*:=\min_{x\in V} \p(x)\ \ {\rm and}\ \ \p^*:=\sum_{x\in V} \p(x).$$ 

Note that for every non-empty subset $A\subsetneq V$, the value of ${(\ph(\pa A)+p(A))}/{\w(A)}$ is a rational number within the interval $[{(\ph_*+\p_*)}/{\w^*}, {(\ph^*+\p^*)}/{\w_*}]$. Furthermore, for two non-empty subsets $A,B\subsetneq V$, if $a:={(\ph(\pa A)+p(A))}/{\w(A)}$ and $b:={(\ph(\pa B)+p(B))}/{\w(B)}$ are distinct, then 
\begin{equation}\label{eq:a}
|a-b|\geq \frac{\ph_*+\p_*}{{\w^*}^2}.
\end{equation}
Based on Algorithm~\ref{alg:daneshgar}, Algorithm~\ref{alg:iso} described below,
finds a minimizer for $\iso_k(T)$.

    \begin{figure*}[t]
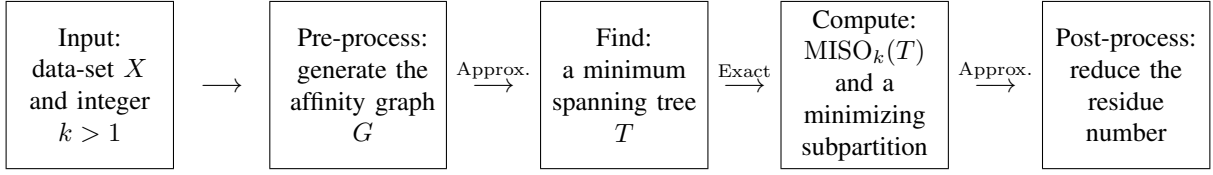

    \centering
    \fbox{\parbox[c][2cm][c]{2cm}{\centering Input:\\ data-set $ X $ and integer $ k>1 $}} $\stackrel{\rm\phantom{Approx.}}{\longrightarrow}$
    \fbox{
\parbox[c][2cm][c]{2cm}{\centering Pre-process:\\ generate the affinity graph $ G $}} $\stackrel{\rm Approx.}{\longrightarrow} $
   \fbox{\parbox[c][2cm][c]{2cm}{\centering Find:\\ a minimum spanning tree $ T $}}  $\stackrel{\rm Exact}{\longrightarrow} $
    \fbox{\parbox[c][2cm][c]{2cm}{\begin{center}
Compute:\\ $ \iso_k(T)$ and a minimizing subpartition
\end{center}  }}
    $\stackrel{\rm Approx.}{\longrightarrow} $
    \fbox{\parbox[c][2cm][c]{2cm}{\centering Post-process:\\ reduce the residue number}}
      \caption{An outline of the main algorithm.}
          \label{fig:outline}
     \end{figure*}

\begin{algorithm}[H]
\caption{Given a weighted tree $(T,\w,\ph,\p)$ and an integer $k$, find a minimizer achieving $\iso_k(T)$.}
\label{alg:iso}
 \begin{algorithmic}
 \STATE Let $\alpha_0\leftarrow\frac{\ph_*+\p_*}{\w^*}$ and $\beta_0\leftarrow\frac{\ph^*+\p^*}{\w_*}$.
 \STATE Let $t\leftarrow \log({2{\w^*}^2(\beta_0-\alpha_0}))-\log ({\ph_*+\p_*})$.
 \STATE Initialize $\alpha\leftarrow \alpha_0$ and $\beta \leftarrow \beta_0$.
 \FOR{$i=1$ \TO $t$}
 \STATE Applying Algorithm~\ref{alg:daneshgar}, decide if $\iso_k(T)\leq \frac{\alpha+\beta}{2}$.
 \IF{$\iso_k(T)\leq \frac{\alpha+\beta}{2}$}
\STATE $\beta\leftarrow \frac{\alpha+\beta}{2}$
\ELSE 
\STATE $\alpha\leftarrow \frac{\alpha+\beta}{2}$
\ENDIF
 \ENDFOR
\STATE Let $\cA$ be the $k$-subpartition output of Algorithm~\ref{alg:daneshgar} for deciding $\iso_k(T)\leq \beta$.
 \STATE \RETURN $\iso_k(T)={\rm cost}(\cA)$ and $\cA$.
 \end{algorithmic}
\end{algorithm}
To prove the correctness of Algorithm~\ref{alg:iso}, note that after the \textbf{for} loop, we obtain an interval $[\alpha,\beta]$, containing rational numbers $\iso_k(T)$ and ${\rm cost(\cA)}$, whose length is equal to 
\[\frac{\beta_0-\alpha_0}{2^t}=\frac{\ph_*+\p_*}{2{\w^*}^2},\]
and consequently, by (\ref{eq:a}), ${\rm cost}(\cA)=\iso_k(T)$.

Finally, the runtime of this algorithm is verified to be in
\begin{align*}
O(nt)&=O\left(n
\left(\log({2{\w^*}^2(\beta_0-\alpha_0}))-\log ({\ph_*+\p_*})\right)\right)\\
&=O(n\log n).
\end{align*}

\end{IEEEproof}

Based on these facts, let us describe the main parts of our proposed clustering algorithm as follows.
(The outline of the algorithm is depicted in Figure~\ref{fig:outline}.)
\begin{enumerate}
\item
Given the data-set of vectors $X$, construct the affinity graph $G$ on $ X $ along with the weights\\ 
$d(x_i,x_j):= \|x_i-x_j\|_{_{2}} $.
\item
Find a minimum spanning tree $T$ of $(G,d)$ and construct a weighted tree $(T,\w,\varphi)$ using the similarity weights as in (\ref{eq:simgraph}).
\item
Apply Algorithm~\ref{alg:iso} to find $\iso_k(T)$ along with a minimizing subpartition $\cA\in \sD_k(X)$.
\item
Use a post-processing algorithm (Algoritm~\ref{alg:post}) to reduce the residue number of $\cA$ and output the optimized clustering $\cA^*$.
\end{enumerate}
\begin{figure}[h]
	\centering
	\includegraphics[scale=.48]{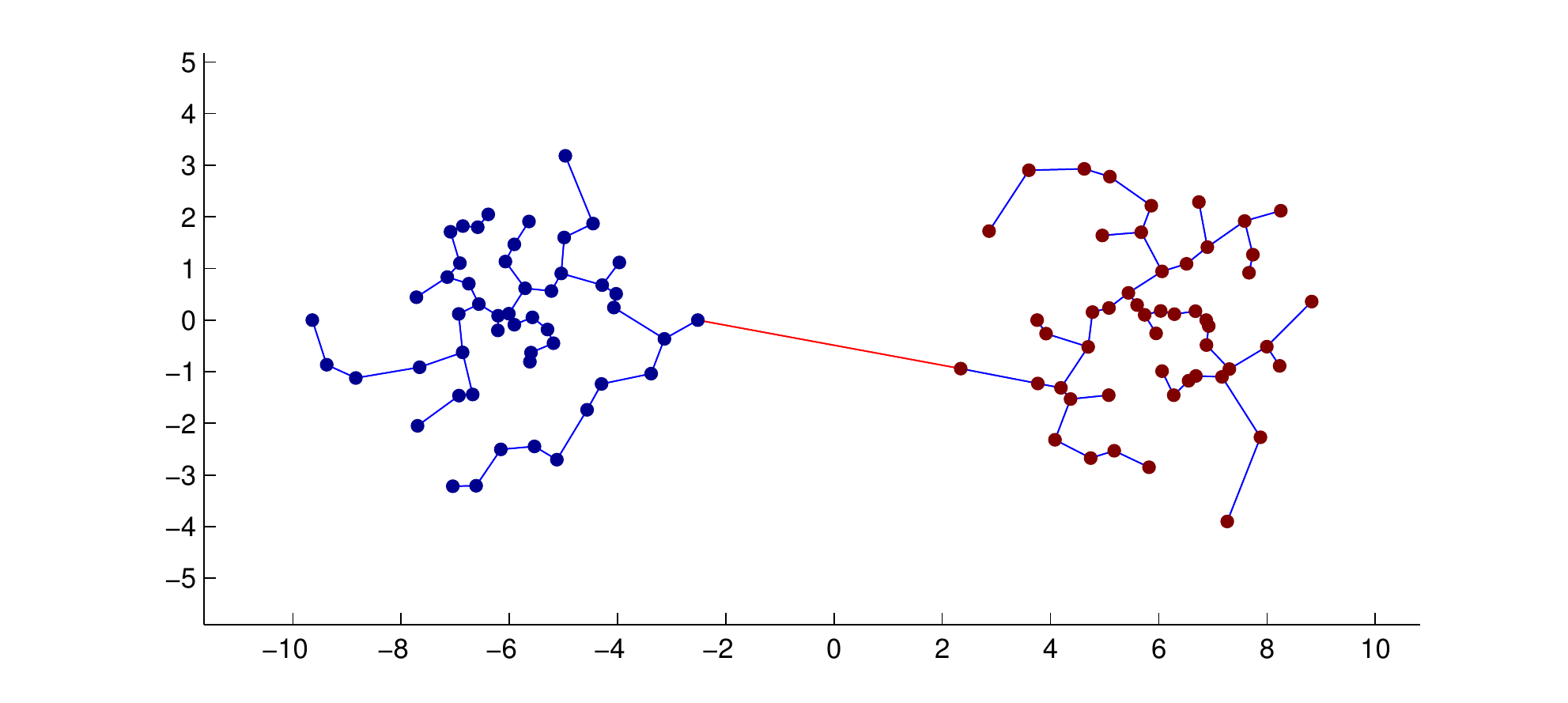}
	\caption{A simple 2-clustering problem and the associated tree and break edge.}
	\label{fig:simpclust}
\end{figure}

The rest of this section is devoted to the post-processing algorithm that tries to reduce the residue number of the subpartition obtained as the output of Algorithm~\ref{alg:iso}. For this it is natural to consider the following 
decision problem.\\ 

\begin{prob}
{MINIMUM RESIDUE NUMBER}
{A weighted tree $T=(V,E,\w,\ph)$ (without potentials) and two integers $k>1, N\geq 0$.}
{Does there exist a minimizing subpartition $\cA\in\sD_k(V)$ achieving $\iso_k(T)$ whose residue number is at most $N$?}
\end{prob}

Unfortunately, the following proposition shows that one may just hope for an approximation of the above problem
since the decision problem is actually $NP$-complete.

\begin{prop}
    \label{pro:residue}
The decision problem {\rm MINIMUM RESIDUE NUMBER} is $NP$-complete in the strong sense for weighted trees.
\end{prop}
\begin{IEEEproof}
See Appendix for a proof.
\end{IEEEproof}

\begin{figure*}[hbt]
\includegraphics[scale=1]{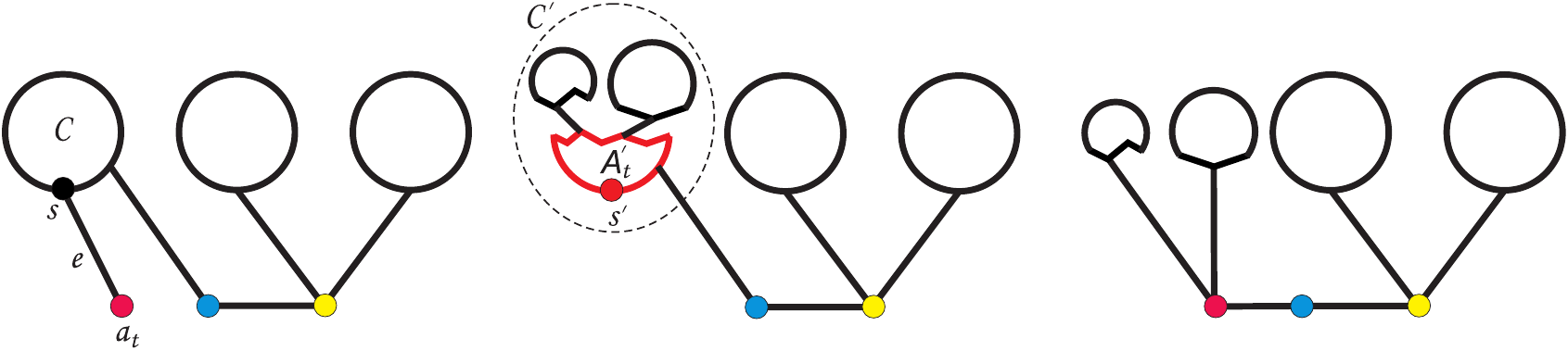}
\caption{A typical scheme of the post-process subroutine.}
\label{fig:scheme}
\end{figure*}

In what follows we propose an approximation scheme for the above problem which will constitute the post-process part of the main algorithm (a schematic general case of this procedure is depicted in Figure~\ref{fig:scheme} which can be helpful in following the details that will follow).

In order to  ensure a high intra-cluster similarity we try to force the induced subgraph on each cluster to form a  connected subgraph. Therefore,  our basic strategy is to look for a minimizing subpartition with a small residue number whose parts  induce connected subgraphs.

To find such a subpartition, with an initial good subpartition in hand, we follow the following post-processing procedure that checks the two following facts,
    \begin{enumerate}
    \item Each part of the subpartition induces connected subgraphs.
    \item No subset of residue elements can be added to any part to make a better subpartition with the same connectivity property.
    \end{enumerate}
    
For this, assume that $\cA=\{A_1,\ldots,A_k\}\in \sD_k(X)$ is the minimizing subpartition, obtained from Algorithm~\ref{alg:iso}. Also, in order to keep the pseudo-code concise we introduce the following terms.
\begin{enumerate}
      \item \textbf{Non-residue vertex:} A vertex in $\cup_{i=1}^k A_i$.
    \item \textbf{Break edge:} An edge in $E(A_i,A_i^c)$ for some $ i $.
    \item \textbf{Residue subtree:} A subtree obtained by removing all break edges from the original tree whose all vertices are residue elements.
     \item \textbf{Start vertex of a residue subtree:} A residue element in the residue subtree which is one end of a break edge.
    \end{enumerate}
Figure~\ref{fig:simpclust} shows a typical pattern of a break edge and the corresponding tree for a simple 
2-clustering problem using the above algorithm.
        
The post-process algorithm can be summarized as follows (see Figure~\ref{fig:scheme}).
\begin{enumerate}
\item
Compute all residue subtrees.
\item 
Contract each $A_i$ with the normalized flow $f_i$ to a single vertex $a_i$ and 
\label{maxflow}
set $a_t:=a_i$ which has the maximum $f_i$.
\item
\label{nextcc}
Let $C$ be a non-flagged residue subtree connected to $a_t$ with start vertex $s$ and break edge $e$.
\item 
Contract edge $e$ to obtain a new residue subtree $C'$ and start vertex $s'$ and update the weights.
\item
Run Algorithm~\ref{alg:post} on $C'$ with the root $s'$ and $N:=\iso(T)$ and let $A'_{t}$ be the output.
\item
If $A'_{t}$ is empty, flag $C$ and goto Step~\ref{nextcc}, otherwise:
	\begin{enumerate}
	\item
	 replace $A_t$ by $A'_t$.
	\item
	update the residue subtree $C$.
	\item
	clear the flags of all residue subtrees.
	\item
	goto Step~\ref{maxflow}.
	\end{enumerate} 
\end{enumerate}

\begin{algorithm}[h]
\caption{Post-process}
\label{alg:post}
\begin{algorithmic}
		\STATE Input a subtree C with the root $s$ and rational number $N$.
		\STATE Order the vertices of $C$ in BFS order as $x_1,x_2,\ldots, x_t=s$.
		\STATE Set $i=1$ and initialize set function $\eta : V(C) \to \mathcal{P}(V(C))$ by $ \eta(x_i) := \{x_i\}$.
		\FOR {$i =1$ \TO $t$}
%			\STATE $v = depthorder(i)$.
			\STATE $u = parent(x_i)$.
			\STATE $e = \{x_i,u\}$.
			\IF{$\p(x_i) - \ph(e) \leq N \  \w(x_i) $}
				\STATE $\p(u) \leftarrow \p(u) + \p(x_i)$.
				\STATE $\w(u) \leftarrow \w(u) + \w(x_i)$.
				\STATE $\eta (u) \leftarrow \eta(u) \cup \eta(x_i)$.
			\ELSE
				\STATE $\p(u) \leftarrow \p(u)  + \ph(e)$.
			\ENDIF
		\ENDFOR
		\RETURN  $\eta(s)$ 	
\end{algorithmic}
\end{algorithm}

In order to prove that the procedure performs correctly, we should prove that searching inside each residue subtree is sufficient to ensure the properties mentioned before. For this, assume that $\cA=\{A_1,\ldots,A_k\}\in \sD_k(X)$ is a minimizing subpartition. Let $C_1$ and $C_2$ be two residue subtrees and for each $i=1,2$, let $S_i\subset C_i$ be a subset connected to $A_1$. Then we have 
\begin{equation}
\label{eq:1}
\frac{\ph(\pa A_1)+\p(A_1)}{\w(A_1)}\leq \iso_k(T).
\end{equation}
Now, if 
\begin{equation}
\label{eq:2}
\forall\ i=1,2,\ \frac{\ph(\pa (A_1\cup S_i))+\p(A_1\cup S_i)}{\w(A_1\cup S_i)}> \iso_k(T),
\end{equation}
then from (\ref{eq:1}) and (\ref{eq:2}), we conclude that 
\[
\frac{\ph(\pa (A_1\cup S_1\cup S_2))+\p(A_1\cup S_1\cup S_2)}{\w(A_1\cup S_1\cup S_2)}> \iso_k(T).\]
This clearly shows that searching inside each residue subtree for a good subset is sufficient to find all good subsets, and hence, the post-process algorithm performs correctly.

%% file: performanceanalysis.tex
\section{Analysis and Experimental Comparison}\label{sec:tests}
In this section we go through the time complexity and performance analysis of our proposed algorithm.

\begin{table*}[t]
	\caption{Performance on UCI database (global scaling): NJW=Ng-Jordan-Weiss \cite{NGJOWE01},	LT=Li-Tian \cite{LITI07}, GS=Grady-Schwartz \cite{GRSCH06}, SM=Shi-Malik \cite{SHIMAL00}, WJHZQ=Wang {\it et. al.} \cite{WA08},  DJS=this paper.}
	\centering
	\begin{tabular}{ |c ||c c c c c c c c c| }
    	\hline
    	
& & & & & $\sigma = 0.09$ & & & &\\ \hline
    	Data set 				& Size & Cluster No. & Dim. &	NJW 	& LT		 	& GS		& SM	&WJHZQ	 & DJS	   \\ \hline %  & DJS*
Wine 					& 178 & 3 & 13 & 0.331461 & 0.286517 & 0.471910 & 0.297753 & 0.325843 & 0.280899  \\ \hline % & 0.280899
Iris						& 150 & 3 & 4   &0.100000 & 0.066667 & 0.333333 & 0.100000 & 0.040000 & 0.040000 \\ \hline %  & 0.040000  
Breast 					& 106 & 6 & 9   &0.632075 & 0.594340 & 0.603774 & 0.698113 & 0.471698 & 0.500000 \\ \hline %  & 0.504762
Segmentation 		& 210 & 7 & 19 &0.561905 & 0.476190 & 0.566667 & 0.371429 & 0.395238 & 0.409524 \\ \hline % & 0.494565
% alpha = 0   => 0.556075
Glass 		 			& 214 & 6 & 10 &0.556075 & 0.457944 & 0.588785 & 0.467290 & 0.658879 & 0.528037 \\ \hline
% alpha = 0   => 0.556075
Average				&       &	&		&	0.4363   & 0.3763    &0.5129    	&0.3869   		&0.3783    &0.3573 			\\ \hline
   \end{tabular}
	\label{tab:uci}
\end{table*}

\begin{table*}[t]
	\caption{Performance on UCI database (local scaling): NJW=Ng-Jordan-Weiss \cite{NGJOWE01},	LT=Li-Tian \cite{LITI07}, GS=Grady-Schwartz \cite{GRSCH06}, SM=Shi-Malik \cite{SHIMAL00}, WJHZQ=Wang {\it et. al.} \cite{WA08},  DJS=this paper.	}
	\centering
	\begin{tabular}{ |c ||c c c c c c c c c| }
    	\hline
    	    	
& & & & & $\nu = 30$ & & & &\\ \hline

    	Data set 				& Size & Cluster No. & Dim. &	NJW 	& LT		 	& GS		& SM	&WJHZQ	 & DJS	    \\ \hline %& DJS*   
Wine 		 			& 178 & 3 & 13 & 0.280899 & 0.280899 & 0.280899 & 0.280899 & 0.308989 & 0.280899   \\ \hline %& 0.269663
Iris 						& 150 & 3 & 4 & 0.086667 & 0.073333 & 0.333333 & 0.100000 & 0.040000 & 0.040000    \\ \hline % & 0.040000
Breast 			 		& 106 & 6 & 9 & 0.622642 & 0.622642 & 0.622642 & 0.641509 & 0.471698 & 0.509434    \\ \hline %&  0.542857
Segmentation 		& 210 & 7 & 19 & 0.404762 & 0.504762 & 0.609524 & 0.404762 & 0.347619 & 0.423810   \\ \hline % &0.304762
Glass 			 		& 214 & 6 & 10 & 0.560748 & 0.495327 & 0.504673 & 0.644860 & 0.658879 &  0.560748  \\ \hline %&0.546729 
Average 					&       &	&		&	0.3911    &  0.3954   &  0.4702     & 0.4144     &0.3654   &  0.3574        \\ \hline  %& 0.3341

   \end{tabular}
	\label{tab:uci2}
\end{table*}

\subsection{Time complexity analysis}
Based on the details presented in the previous section the algorithm consists of three phases:
\begin{itemize}
\item PHASE I: A pre-processing phase where an affinity graph is constructed from the input data and a minimum spanning tree of the graph is obtained.
\item  PHASE II: A tree-partitioning phase where the minimum spanning tree is sub-partitioned according to an isoperimetry criteria (Algorithm~\ref{alg:iso}).
\item PHASE III: A post-processing phase where residue subtrees are reprocessed in order to find a minimizing subpartition with the minimal residue number.
\end{itemize}
In what follows we elaborate on estimating the time complexity of each phase. We denote the time complexity function of the algorithm with $t(n)$ where $n$ is the size of the data set, where
\begin{equation}
t(n) = t_1(n) + t_2(n) + t_3(n),
\end{equation}
in which $t_i(n)$ denotes the time complexity of the $i$'th phase.
The following analysis will show that $t(n) \in O(n^2)$ in the worst case for the global scaling scenario with a post-processing,
where in the local scaling setting it is observed that the basic algorithm operates in $O(n \log n)$ time.

\subsubsection{{\rm PHASE I} }
It is supposed that the $i$'th object of interest is given in a vector representation $F_{[d \times 1]}^{i}$ where each element of the feature vector is a real number ($d$ is a fixed integer). Also, it is presumed that a similarity function $S: \mathbb{R}^d \times \mathbb{R}^d \rightarrow \mathbb{R} $ is given with computational complexity $O(t_S(d)) = O(c)$ for a constant $c$.

In construction of a {\it global scale} affinity graph one needs $ \binom{n}{2} \in O(n^2)$ times computation of the similarity function. On the other hand, in a  {\it local scale} affinity graph (e.g. see \cite{ZEPE04}) it is enough to focus on a fixed proximity of each vertex (e.g. as is the case in image segmentation application) and consequently,
one may use any technique for nearest neighbour search problem as space partitioning, locality sensitive hashing, approximate nearest neighbour or other well-known methods  to obtain the affinity graph of the input vectors in 
sub-polynomial time. Assuming that the input vectors belong to a space with a Minkowski metric (which is usually the case in real applications) one may use $\epsilon-$approximate nearest neighbour
%($\epsilon-$ANN for short)
%(e.g. see \cite{Arya94anoptimal})
 to find a fixed number $\nu$ of neighbours for each object that results in a graph with size $|E| \in O(n)$ which is constructed in time $ O( n \log n)$ for a fixed $\epsilon$.

Given a graph of size  $|E| \in O(n)$ one may easily find a minimum spanning tree using well-known algorithms 
%(e.g.  Kruskal's algorithm \cite{citeulike:4031585,Pettie:2002,Pettie:2008:RMS:1328911.1328916})
 in time at most $O(n \log n)$. 

Hence, in a local scaling model we may assume that $t_1(n) \in O(n \log n)$ and in the global scaling model 
we have $t_1(n) \in O(n^2)$ in the worst case.

\subsubsection{{\rm PHASE II}}
Time complexity of Algorithm~\ref{alg:daneshgar} was shown to be linear in \cite{DAM10}.  
We verified within the proof of Theorem~\ref{thm:effcomputisotree}, that the runtime of Algorithm~\ref{alg:iso} lies in $O(n\log n)$. Thus, with the notations in the proof of Theorem~\ref{thm:effcomputisotree}, we have
\begin{align*}
t_2(n)&\in O\left(n
\left(\log({2{\w^*}^2(\beta_0-\alpha_0}))-\log ({\ph_*+\p_*})\right)\right)\\
&=O(n\log n).
\end{align*}
%
%Let us  define,
%
%\begin{equation*}
%\begin{split}
%\bar{\ph}:=  \max_{e \in E} \ph(e), \hspace{1cm}  \bar{\w}:= \max_{x \in V} \w(x).
%\end{split}
%\end{equation*}
%
%Now, applying results of \cite{DAM10} the time complexity of tree partitioning phase can be written as,
%
%\begin{equation}
%\label{equitnum}
%O(n\lceil \log (\frac{\ph^* \w^*}{\ph_* \w_* 2\epsilon}- \frac{1}{\epsilon} ) \rceil)).
%\end{equation}
%
%However, on the other hand, we have 
%
%\begin{equation*}
%\frac{\ph^* \w^*}{\ph_* \w_* 2\epsilon}- \frac{1}{\epsilon}  < \frac{n^2 \bar{\ph} \bar{\w} }{\ph_* \w_* 2\epsilon}- \frac{1}{\epsilon} <  c' \frac{n^2 }{2\epsilon}- \frac{1}{\epsilon}
%\end{equation*}
%
%where $c'$ is defined as the following ratio,
%
%\begin{equation}
%c' := \frac{\bar{\ph} \bar{\w}}{\ph_* \w_*}.
%\end{equation}
%
%Also, defining the vertex weights as the sum of weights on the incident edges, one may compute
%
%\begin{equation*}
%c' = \frac{\bar{\ph}}{\ph_*} . \frac{\bar{\w}}{\w_*} \leq \frac{\bar{\ph}}{\ph_*} . \frac{n\bar{\ph}}{\ph_*} = n ({\bar{\ph}}{\ph_*})^2.
%\end{equation*}
%
%Finally, fixing precision bounds for the edge weights and $\epsilon$ we have,
%
%$$
%\begin{array}{ll}
%t_2(n) & \in O(n \lceil \log (c' \frac{n^2 }{ 2\epsilon}- \frac{1}{\epsilon} ) \rceil )\\
%&= O(-n \log \epsilon +3n \log n)=O(n \log n).
%\end{array}
%$$

\subsubsection{{\rm PHASE III}}

By definitions the worst case time complexity of the post-processing algorithm is bounded by $O(n^2)$, since one may think of a case where all connected components of the subgraph are of order one.
However, it should be noted that in our real experiments the generic cases were observed  to be far from the worst case.

\subsection{Experimental results}
In this section we provide our experimental comparison of proposed algorithm with some similar algorithms on some well-known clustering datasets. In our experiments the edge weights are assumed to be equal to
\begin{equation}
\ph(uv) = \exp (-\frac{S(u,v)}{\sigma})
\end{equation}
following the convention in  clustering literature, where $\sigma$ is the scaling  parameter.
Also, hereafter, $\nu$ stands for the neighbourhood parameter for local scaling.

\subsubsection{{\rm UCI} benchmarks}
Table~\ref{tab:uci} reports the outcome of the performance analysis of four algorithms on UCI machine-learning benchmark repository \cite{UCI}. Each number in the table represents the misclassification rate (i.e. the ratio of incorrect labellings to the total number of objects) for the corresponding algorithm. For the algorithms that 
do not get the number of clusters as a part of the input a precision threshold is set and the algorithm is terminated 
after that stage. We should also report that the GS algorithm just terminated on the Iris data set with a $2$-clustering and this is the main reason for this algorithm relatively high misclassification rate in this case.

\subsubsection{Some hard instances}
\begin{figure}[ht]
        \caption{The first column contains a set of hard problems created artificially and the second column consists of four cases chosen from \protect\url{http://www.vision.caltech.edu/lihi/Demos/SelfTuningClustering.html} where the local scaling parameter is set to $\nu=7$ ($\sigma = 0.1$ for all problems).}
        \vspace{10pt}
        \centering
        \begin{tabular}{|c|c|}
        \hline
        \includegraphics[scale=.22]{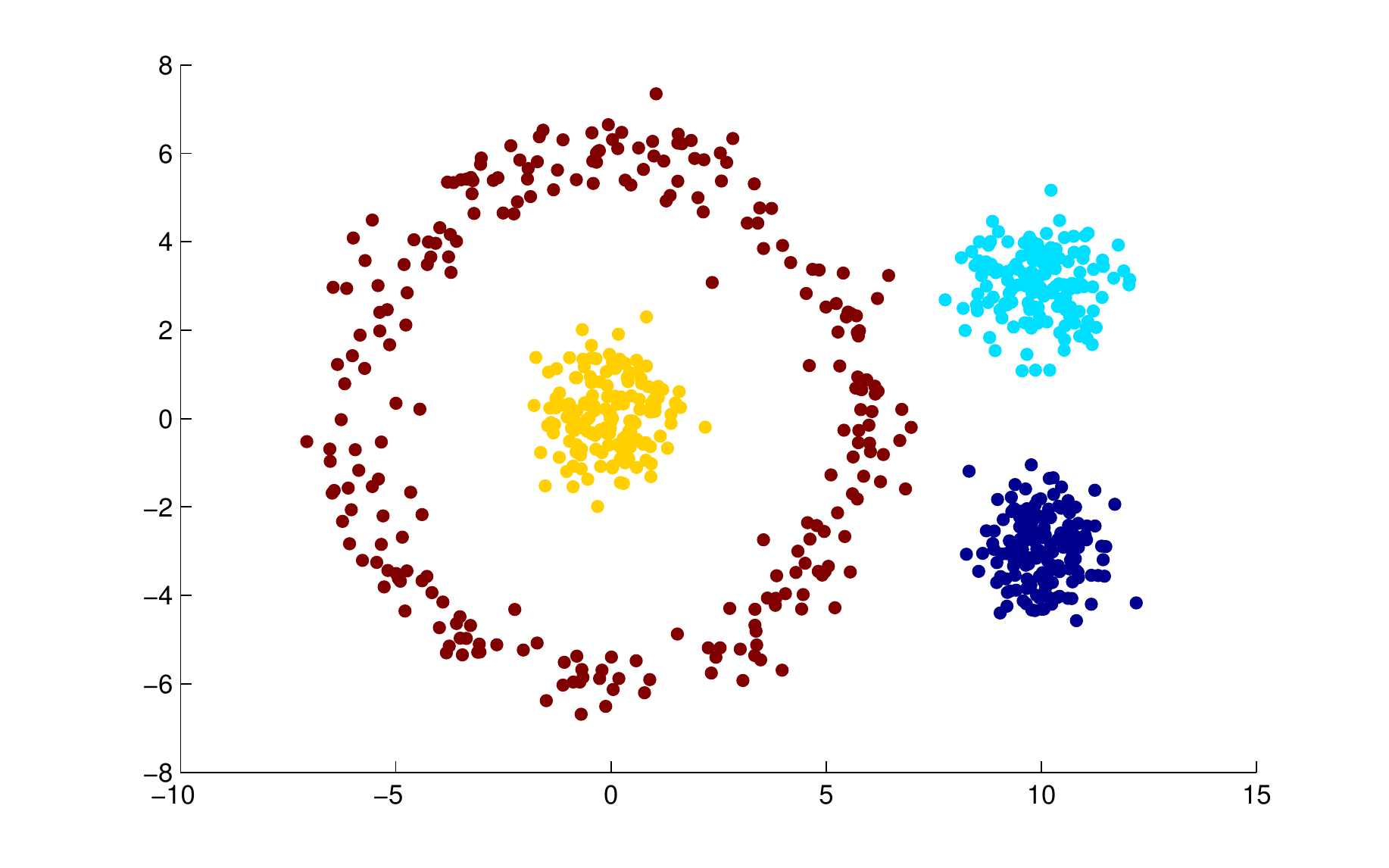}&\includegraphics[scale=.22]{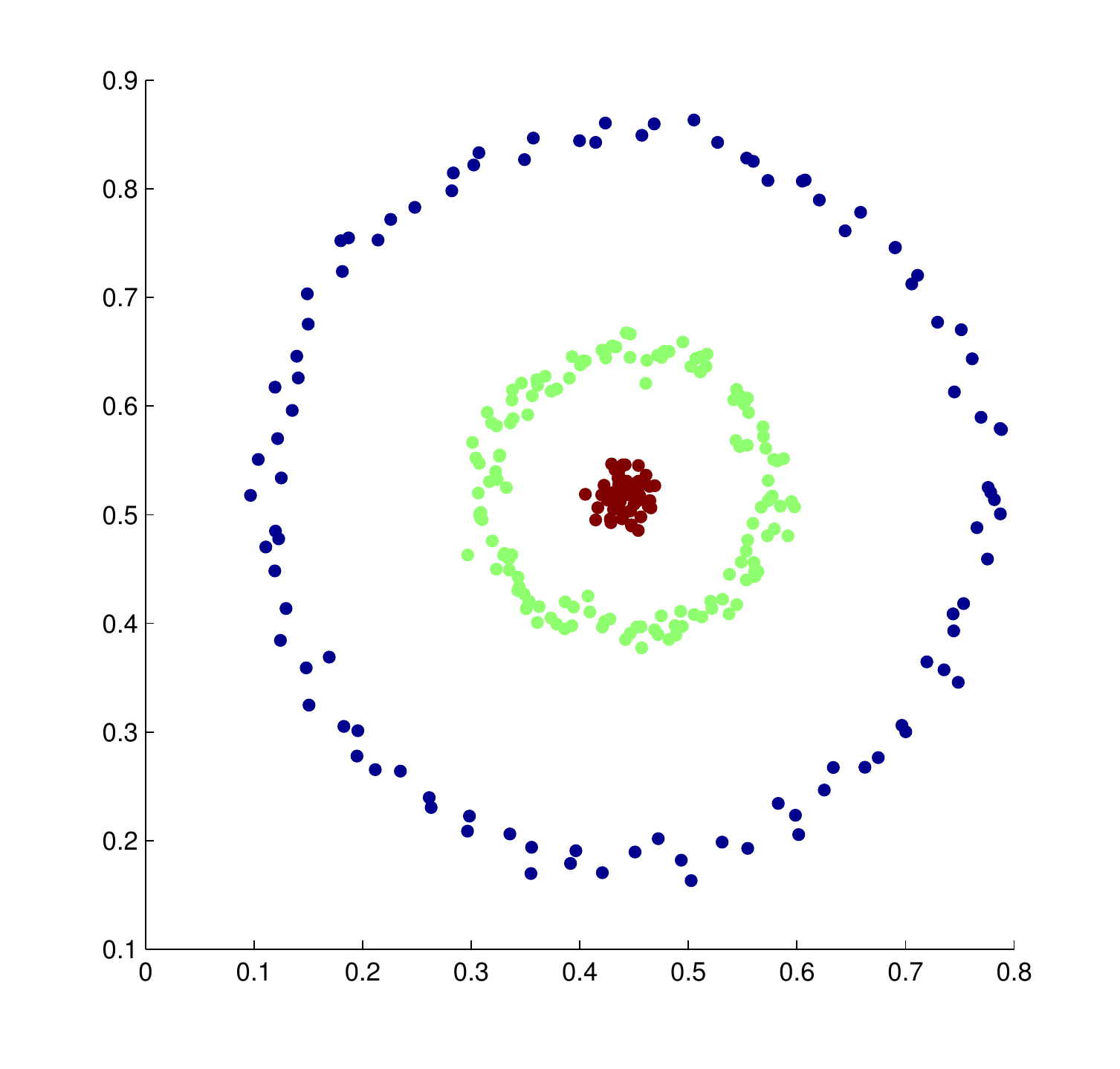}\\ \hline
        \newline
        \includegraphics[scale=.22]{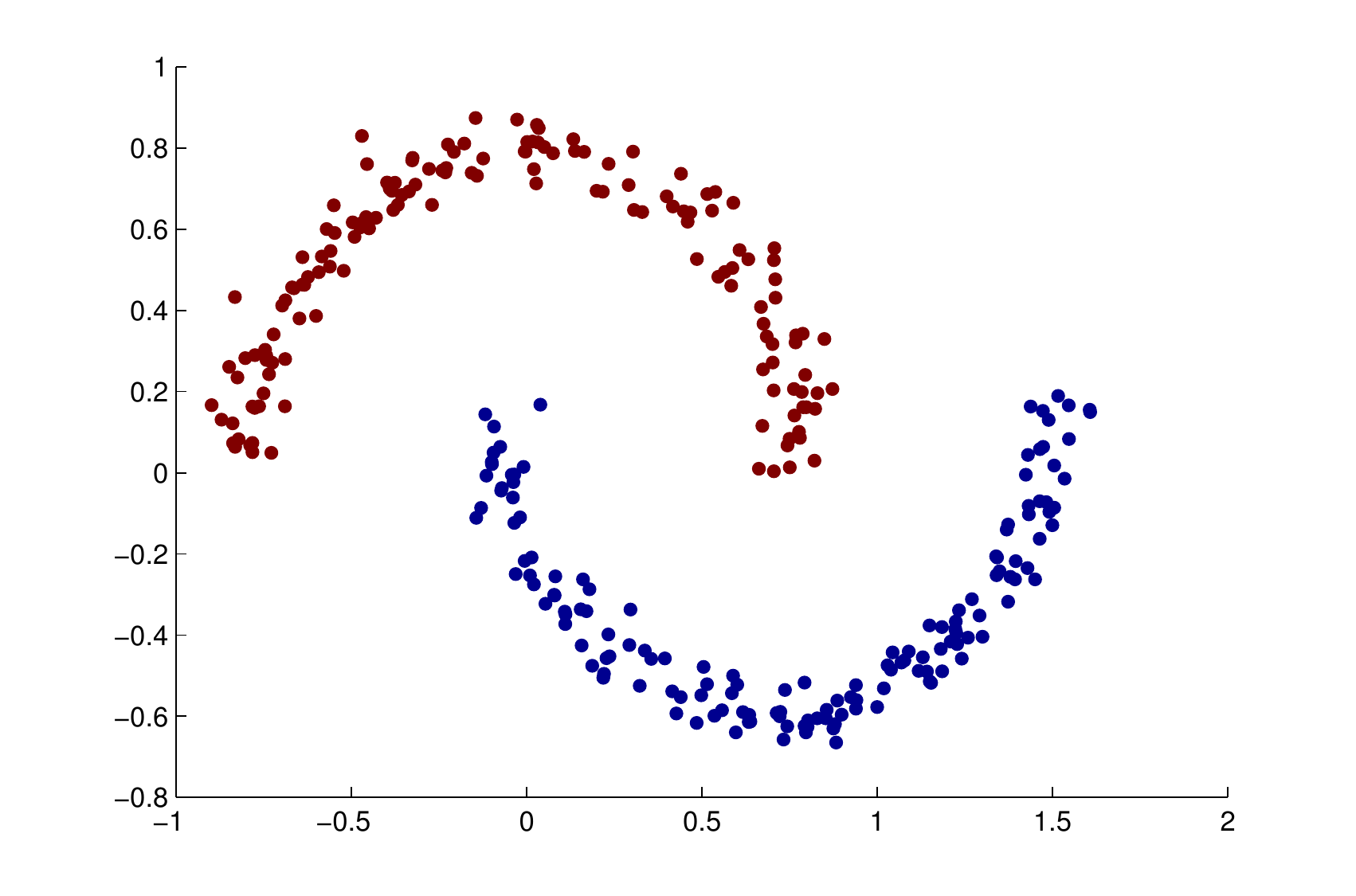}&\includegraphics[scale=.22]{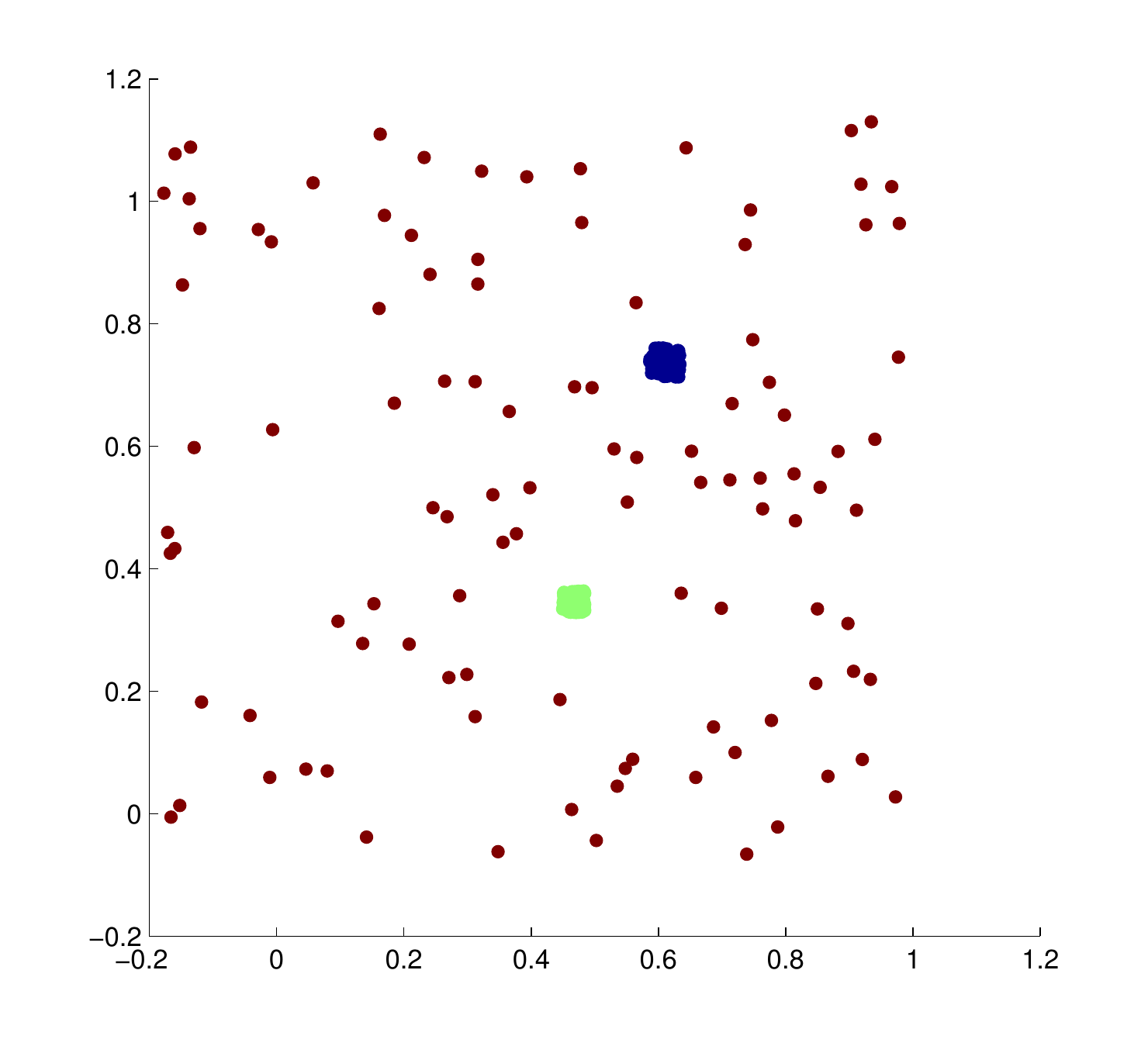}\\ \hline
        \newline
        \includegraphics[scale=.22]{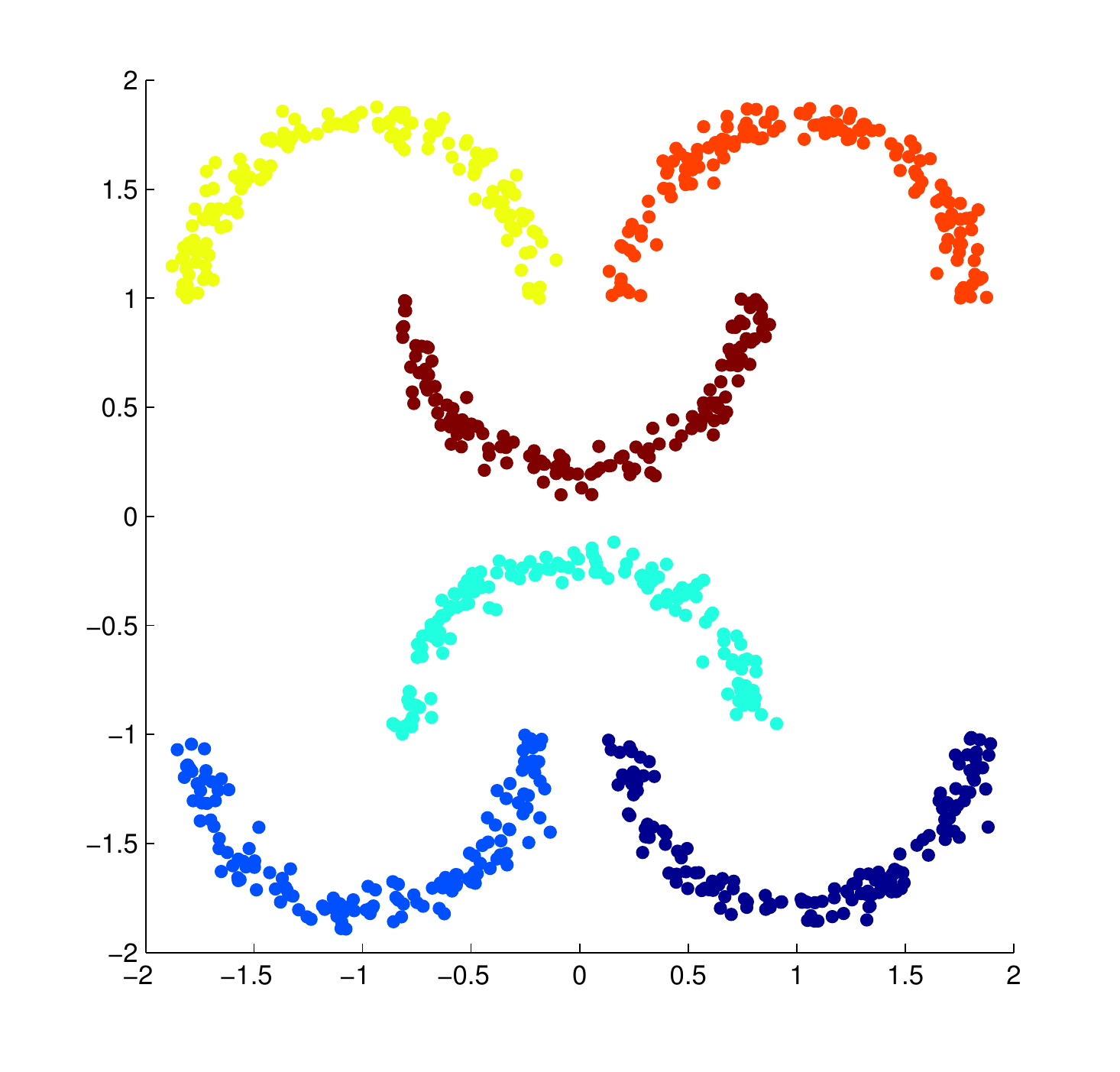}&\includegraphics[scale=.22]{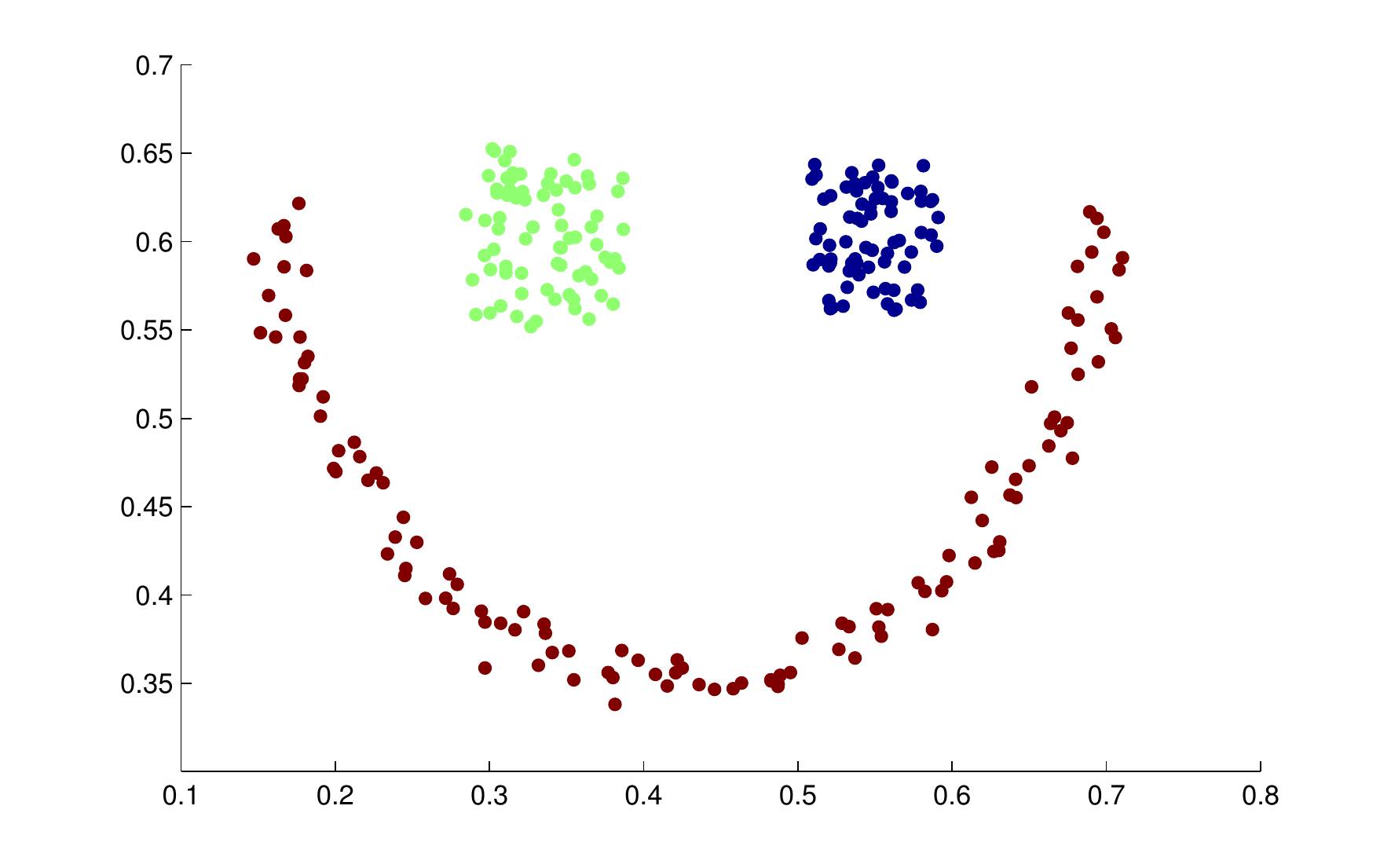}\\ \hline
        \newline
        \includegraphics[scale=.22]{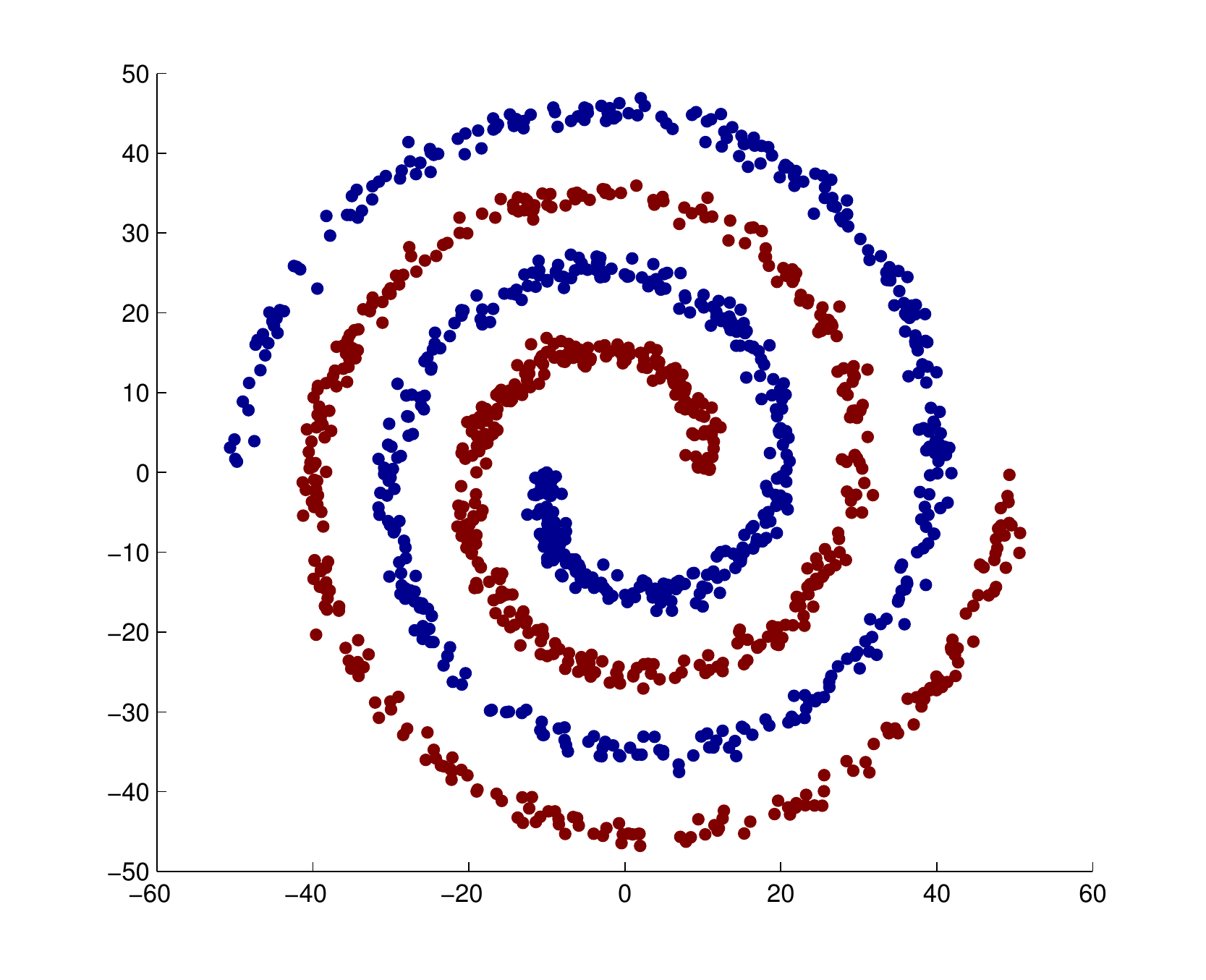}&\includegraphics[scale=.22]{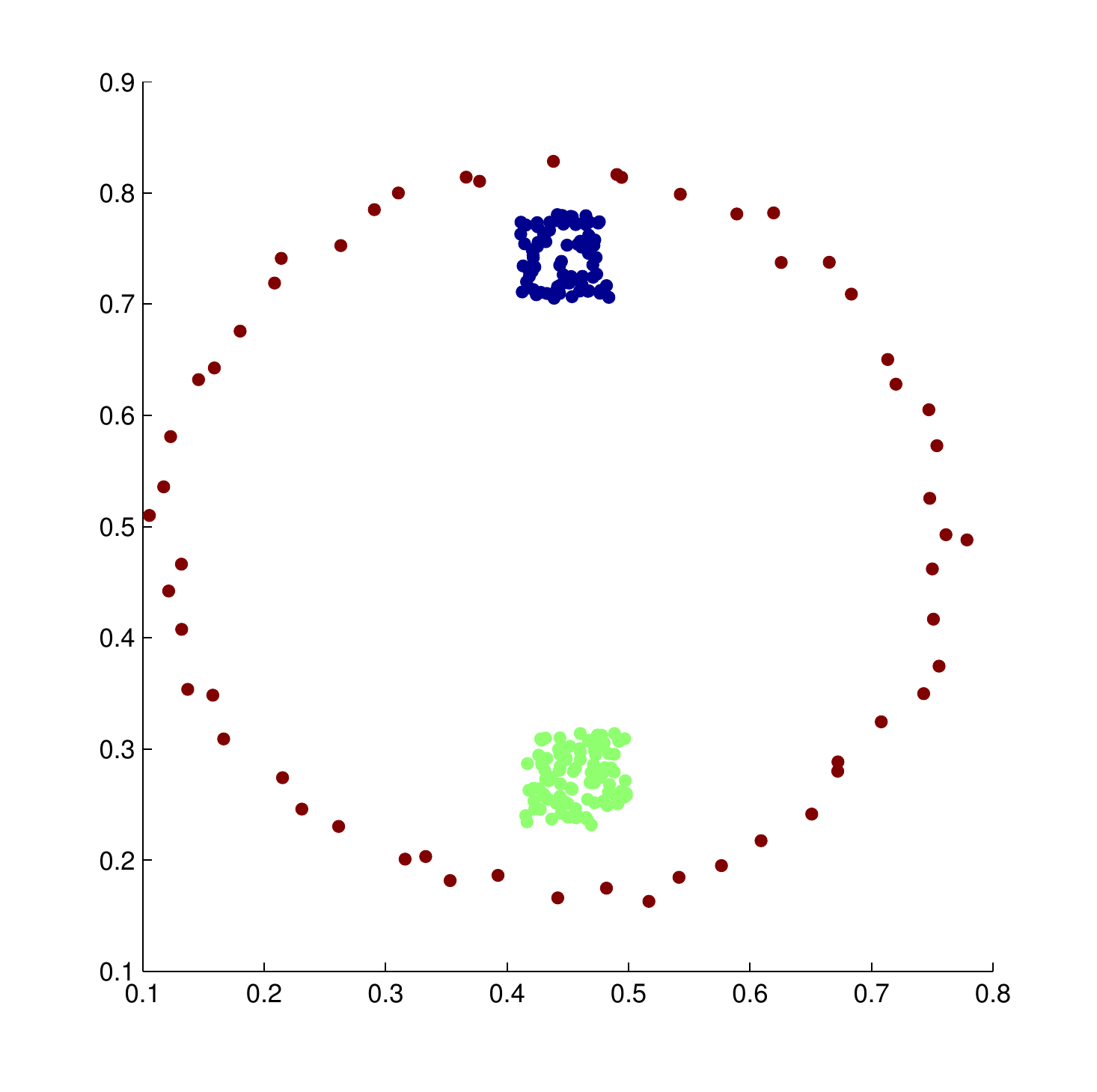}\\ \hline
        \end{tabular}
        \label{fig:pointclus}
	\end{figure}

We have also considered the performance of our algorithm on a couple of hard artificial clustering problems as is depicted  in Figure~\ref{fig:pointclus}. As it is clear from the results the algorithm has been successful enough to extract the correct expected clusters.

%% file: outlierprofile.tex
\section{The outlier profile of a data-set}\label{sec:outlierprofile}
In this section we try to show how our new extended model can be used to formalize the concept of an outlier set.
First of all, it should be noted that the concept of an {\it outlier } is not easy to formalize since it certainly 
depends on the data scaling factors. Although, there has been very little on the formalization of this concept  (e.g. see \cite{vempala} as a very exceptional nice reference), we believe that one should try to define the concept of an 
{\it outlier profile} of a data set (to be made precise later) than trying to define the concept of {\it an outlier set}, since depending on ones precision about the concept of {\it being far}, one may come to very different conclusions about what an outlier is. Hence, in what follows we use the flexibility of {\it potentials}, already introduced in our model, to 
simulate this precision analysis, and consequently, we will be able to define the concept of an outlier set with respect to a precision parameter $\alpha$. Therefore, using this parametrization one comes to a varying 
family of sets that will be called the {\it outlier profile} of the data set. Using this we will also try to extract 
the best candidate for an {\it outlier set} and we will propose a criteria for extracting such a set while we will also provide an approximation algorithm to approach to a solution  for the problem. We also provide experimental results that can serve as evidence to the correctness and applicability of our approach. 

We would like to note that this section is just a starter to these ideas and we believe there is a lot more that should be investigated theoretically and experimentally. \\

\begin{defin}{
Let $G=(X,E,\w,\ph,\p)$ be a weighted graph with potential. Then given $k \in \mathbb{N}$ and  $\alpha \in \mathbb{R}^+$, for any $\cA\in \sD_k(X)$ define 
\begin{equation}\label{eq:costalpha}
{\rm cost}_{k,\alpha}(\cA):= \max_{1\leq i\leq k} \frac{\ph(\pa A_i)+\alpha\ \p(A_i)}{\w(A_i)},
\end{equation}
and
\[\iso_{k,\alpha}(G):= \min_{\cA\in \sD_k(X)} {\rm cost}_{k,\alpha}(\cA).\]
A subset $A \subset V$
is said to be a $(k,\alpha)$-outlier  of $G$ if for any $\beta \geq \alpha$, the subset $A$ does not intersect any minimizer of $\iso_{k,\beta}(G)$. It is clear by definition that any subset of a $(k,\alpha)$-outlier of $G$ 
is also a $(k,\alpha)$-outlier of $G$. Hence, we define the {\it $(k,\alpha)$-outlier set}  of $G$, $O_{k,\alpha}(G)$, to be the
union of all $(k,\alpha)$-outliers of $G$ which is itself a maximal $(k,\alpha)$-outlier of $G$.
}\end{defin}
Note that the concept of an $(k,\alpha)$-outlier also depends on the potential function $\p$. Also,
Now, as a direct consequence of definition we have,

\begin{prop}
Given positive real numbers $\alpha \leq \beta$ and a weighted graph $G=(X,E,\w,\ph,\p)$, then
$O_{k,\alpha}(G) \subseteq O_{k,\beta}(G).$
\end{prop}

In order to match the definition with our intuition about outlier sets we should relate the potential function to the distance (i.e. the inverse of the similarity) function, and for this we adopt the special potential function for which the potential at each vertex is the mean of the distance of the vertex to the rest of the vertices.
\[p(x):=\frac{1}{n} \sum_{y\in X} \|x-y\|_{_{2}}.\]

\begin{figure}[H]
        \caption{\label{fig:alphalevels}The increase of residues as $\alpha$ increases (residue vertices are 
        depicted in black for  $\alpha = 0,0.6,0.8,2,5,50$).}           
        \centering
       \vspace{10pt}
        \begin{tabular}{c}
        \hline
        \includegraphics[scale=.45]{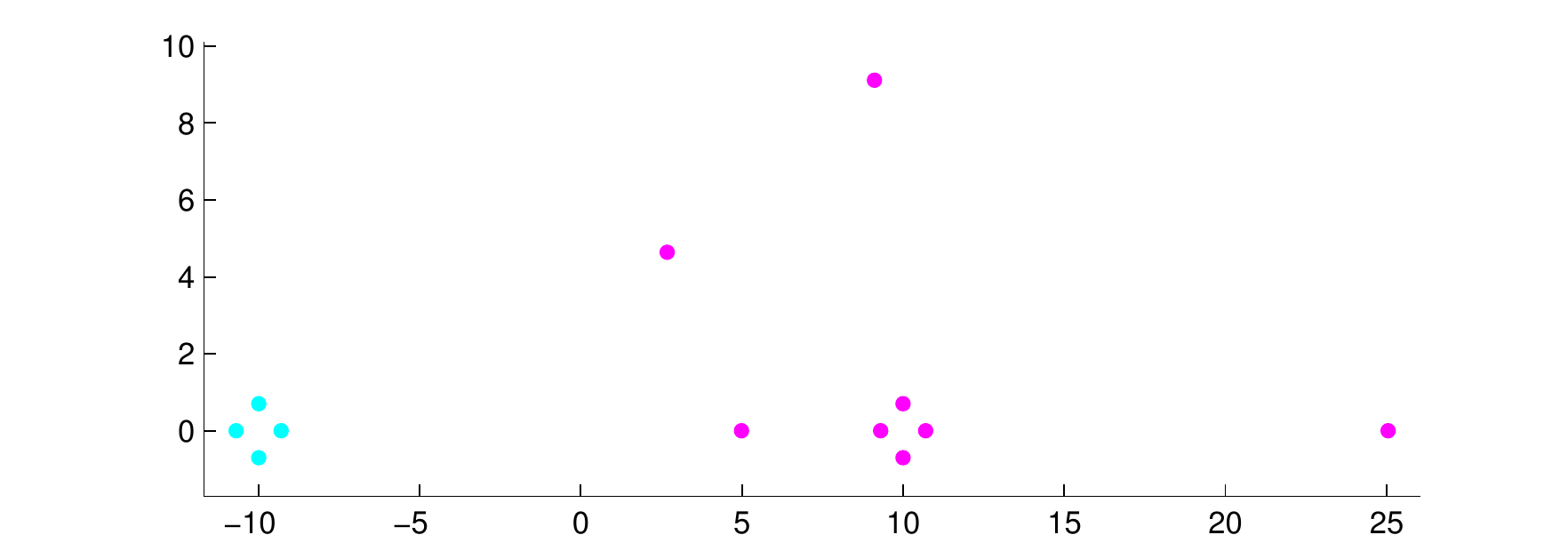}\\
        \includegraphics[scale=.45]{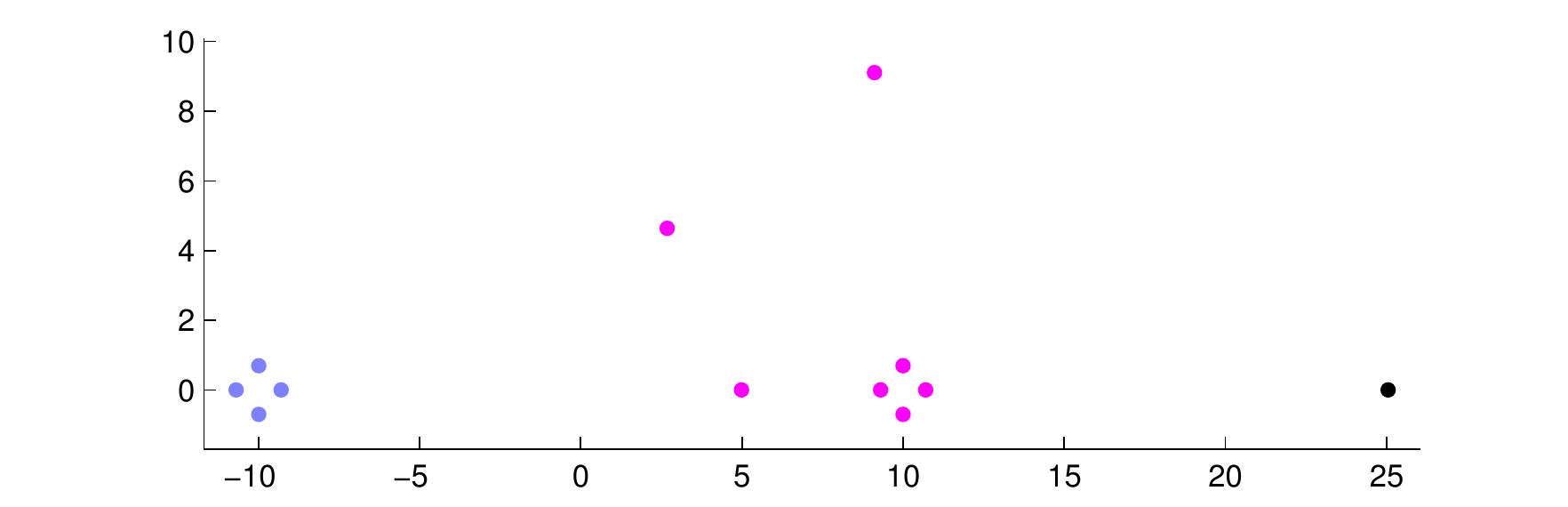}\\
        \includegraphics[scale=.45]{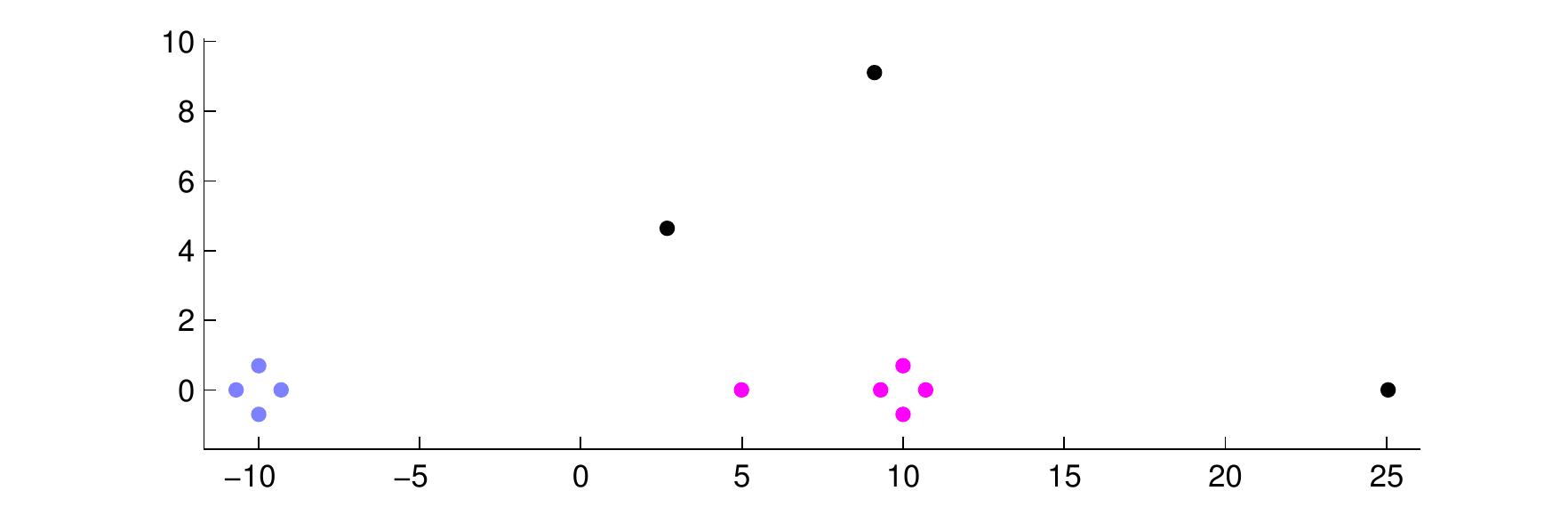}\\
        \includegraphics[scale=.45]{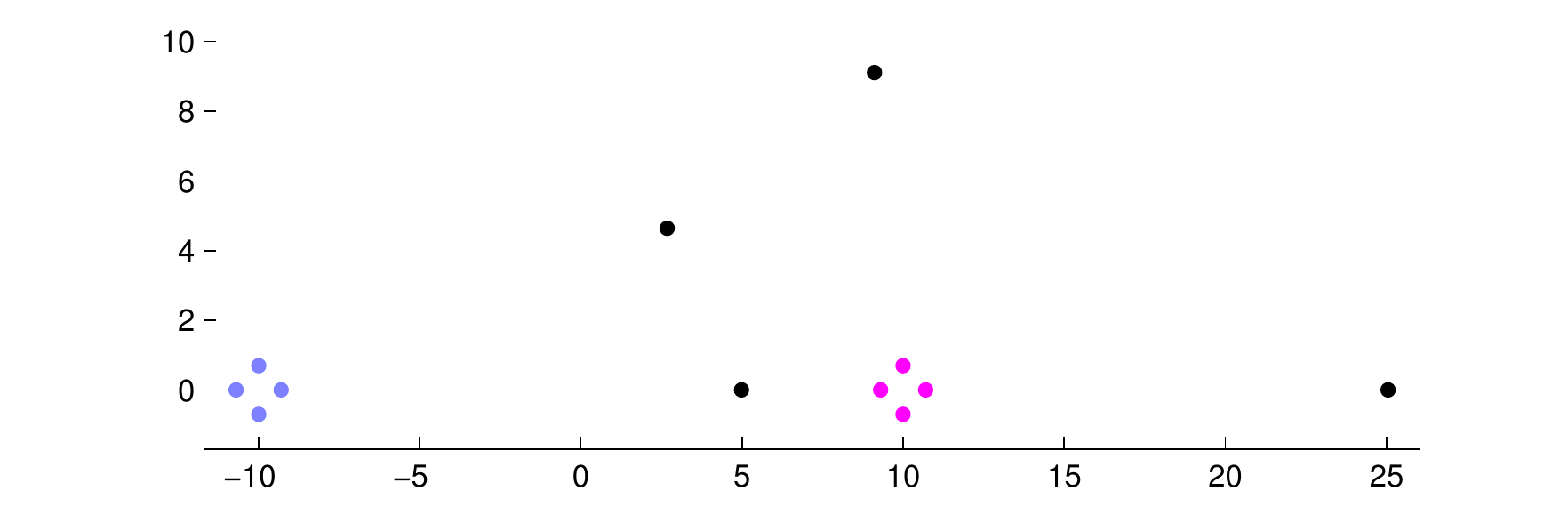}\\ 
        \includegraphics[scale=.45]{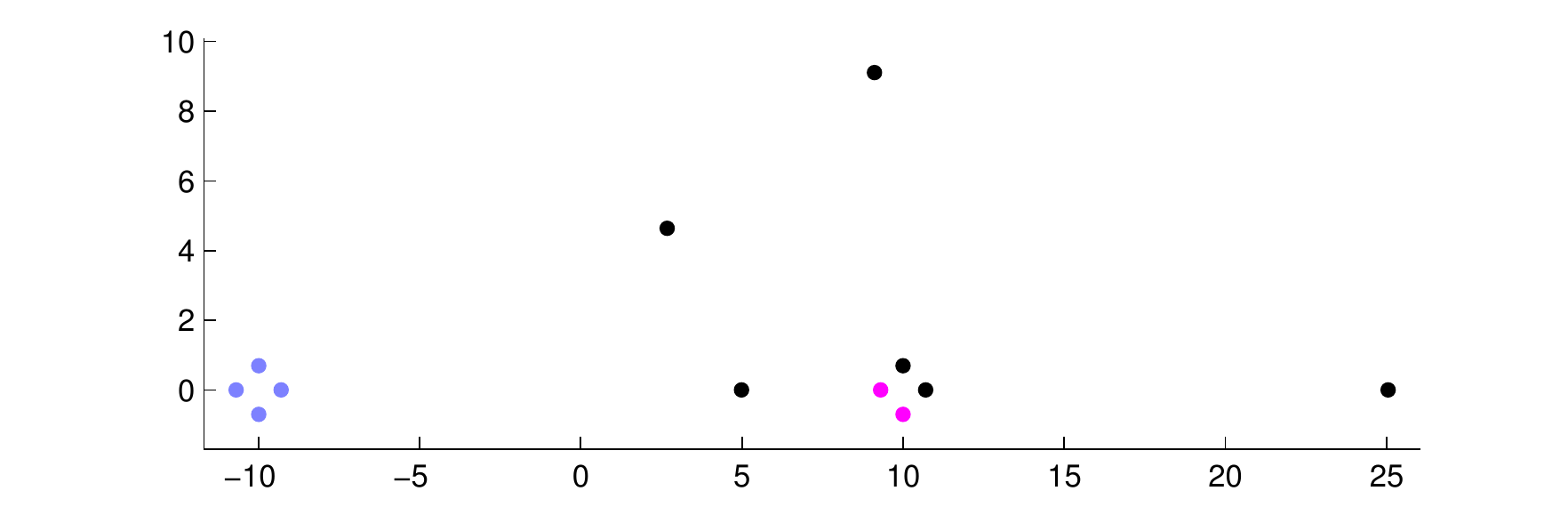}\\
        \includegraphics[scale=.45]{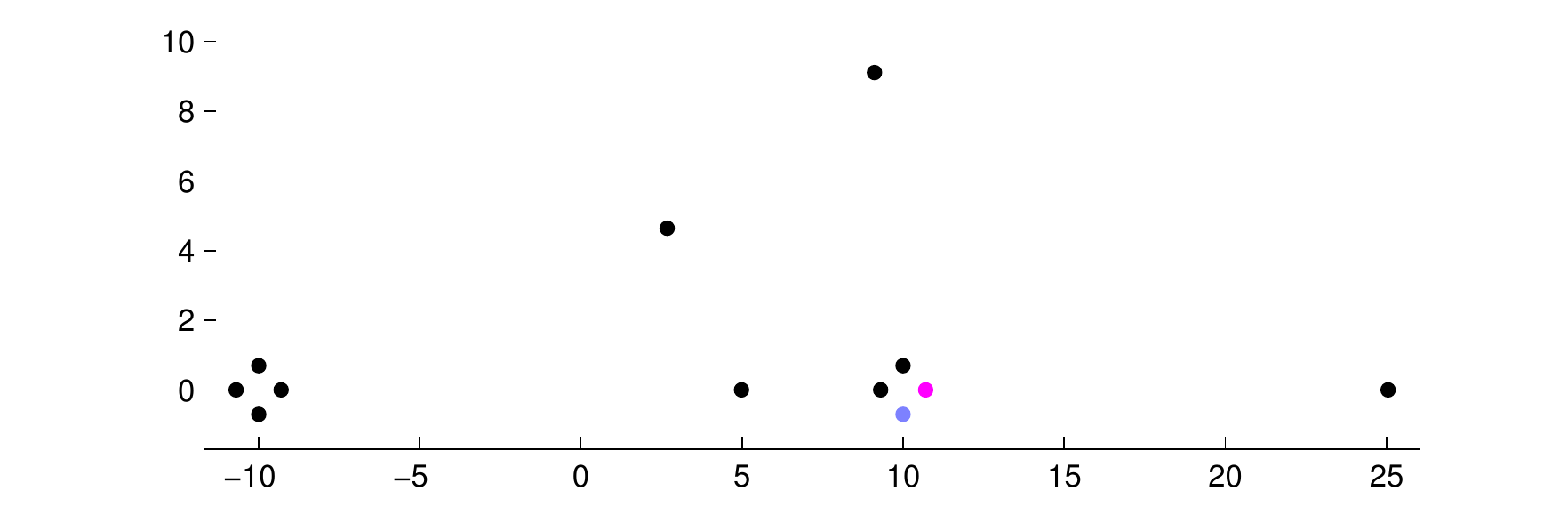}\\
        \newline 
        \end{tabular}
\end{figure}

To get a feeling about how this definition works one may refer to the artificial example depicted in Figure~\ref{fig:alphalevels} in which one can see that the residue of the minimizing subpartition increases as 
a consequence of an increase in the potential function (i.e. an increase of $\alpha$).

\begin{figure*}[ht]
        \caption{\label{fig:hspectrum}An approximation of the outlier profile of the graph depicted in Figure~\ref{fig:alphalevels}.}
        \centering
        \begin{tabular}{c}
        \hline
        \includegraphics[scale=.85]{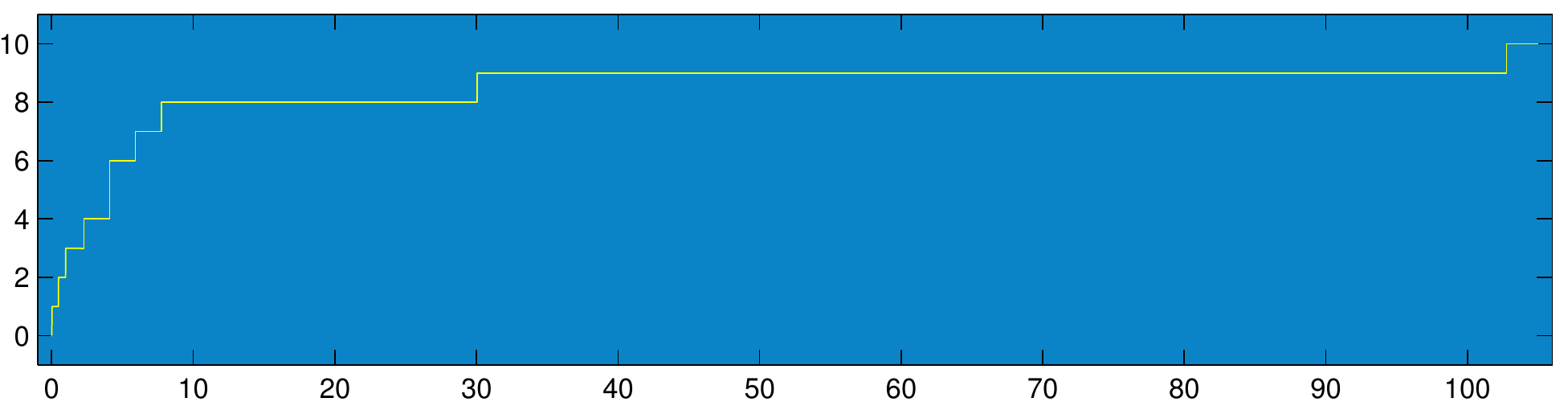}\\
        \newline 
        \end{tabular}
\end{figure*}

Now, the main problem is how one may approximately compute the outlier profile of a best candidate for an outlier set based on the above mentioned definition. For this, consider Figure~\ref{fig:hspectrum} that shows the way of increasing of the residue number as increasing of $\alpha$ for the graph of Figure~\ref{fig:alphalevels}. Hereafter, this function is called the {\it outlier profile} of the graph.

Now, our main problem is to extract the best candidate for an outlier set. For this, one may note that a possible natural criteria for this is a relatively long delay in a constant value in the outlier profile which is also close 
to the origin. As it is clear, the two intuitive conditions namely, being close to the origin (i.e. small $\alpha$)
and having a long stay in a constant value, are contradictory to each other and to set a criterion one needs an scaling factor. Hence we let the parameter $\sigma_s$ to be the desired scaling factor and define

\begin{equation}
\label{eq:spectmeasure}
sm(I_i) = \exp(\frac{-\min(I_i)}{\sigma_s})-\exp(\frac{-(\max(I_i)}{\sigma_s}),
\end{equation}
in which $I_i=[\min(I_i),\max(I_i)]$ is the $i$'th interval on which the number of residues remains constant.
Hence, we let the interval $I_*$ be the interval on which $sm(I_i)$ is maximized and choose this interval as the 
space  in which the best $\alpha$ lives. Following this idea we define 
\begin{equation}
\alpha_* = \min (I_*).
\end{equation}

\begin{algorithm}[ht]
\caption{HeuristicSearch}
\label{alg:spectheu}
\begin{algorithmic}
		\REQUIRE BPs($1 ... n-k$) = [$\infty, -\infty$] , $a<b$, $n_a < n_b$
		\STATE Input $BPs,[a,b],[n_a,n_b],T,P,\epsilon$
		
		\IF {$b-a \leq \epsilon$}
			\RETURN BPs
		\ELSE
			\STATE $\alpha = \frac{a+b}{2}$
			\STATE subpartition = \COMMENT {The output of Algorithm~\ref{alg:iso}}
			\STATE  $resno$ =ResidueNumber(subpartition)
			\IF {$resno \leq n_a$ }
				\STATE Update all BPs from index $resno+1$ to $n_a$
				\STATE BPs = HeuristicSearch($BPs,[\alpha ,b],[resno,n_b],T,P,\epsilon$)
			\ELSIF {resno $\geq n_b$}
				\STATE Update all BPs from index $n_b+1$ to $resno$
				\STATE BPs = HeuristicSearch($BPs,[a,\alpha ],[n_a,n_b],T,P,\epsilon$)
			\ELSE 				
				\STATE Update BPs with index resno
				\IF {resno $> n_a$}
					\STATE BPs = HeuristicSearch($BPs,[a,\alpha ],[n_a,resno],T,P,\epsilon$)
				\ENDIF
				\IF {resno $< n_b$}
					\STATE BPs = HeuristicSearch($BPs,[\alpha,b],[resno,n_b],T,P,\epsilon$)
				\ENDIF
			\ENDIF
		\ENDIF	
\end{algorithmic}
\end{algorithm}

\begin{figure}[ht]
        \caption{\label{fig:hardoutlier2}A problem with local affinity parameter $\nu=20$ and  $\sigma_s = 0.5$ .
        First row contains the outcome of WJHZQ \cite{WA08} Algorithm. Second row contains the outcome of
        our algorithm with $\alpha_* = 0.24$. }
        \centering
        \begin{tabular}{c}
        \includegraphics[scale=.6]{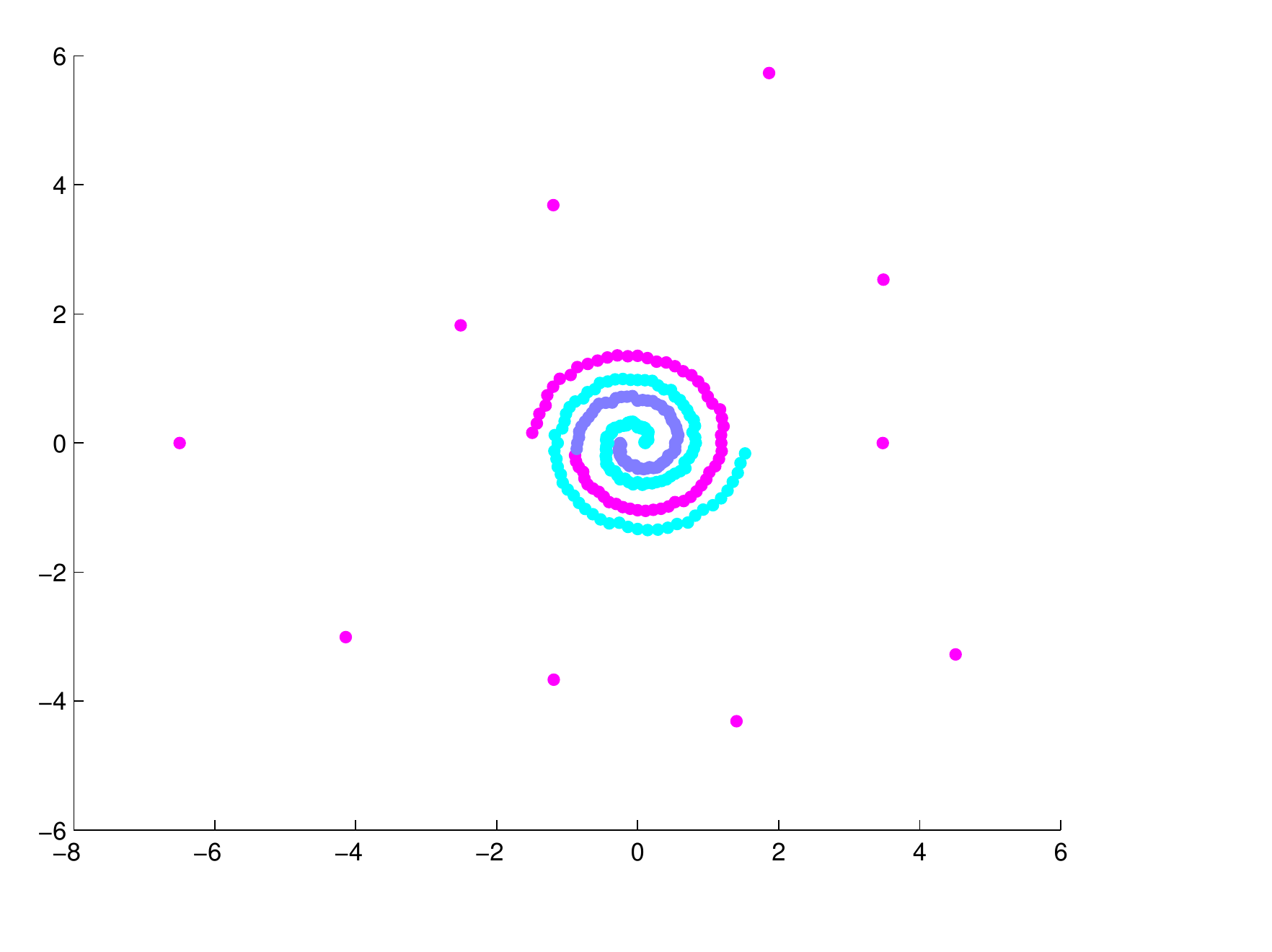}\\
        \includegraphics[scale=.6]{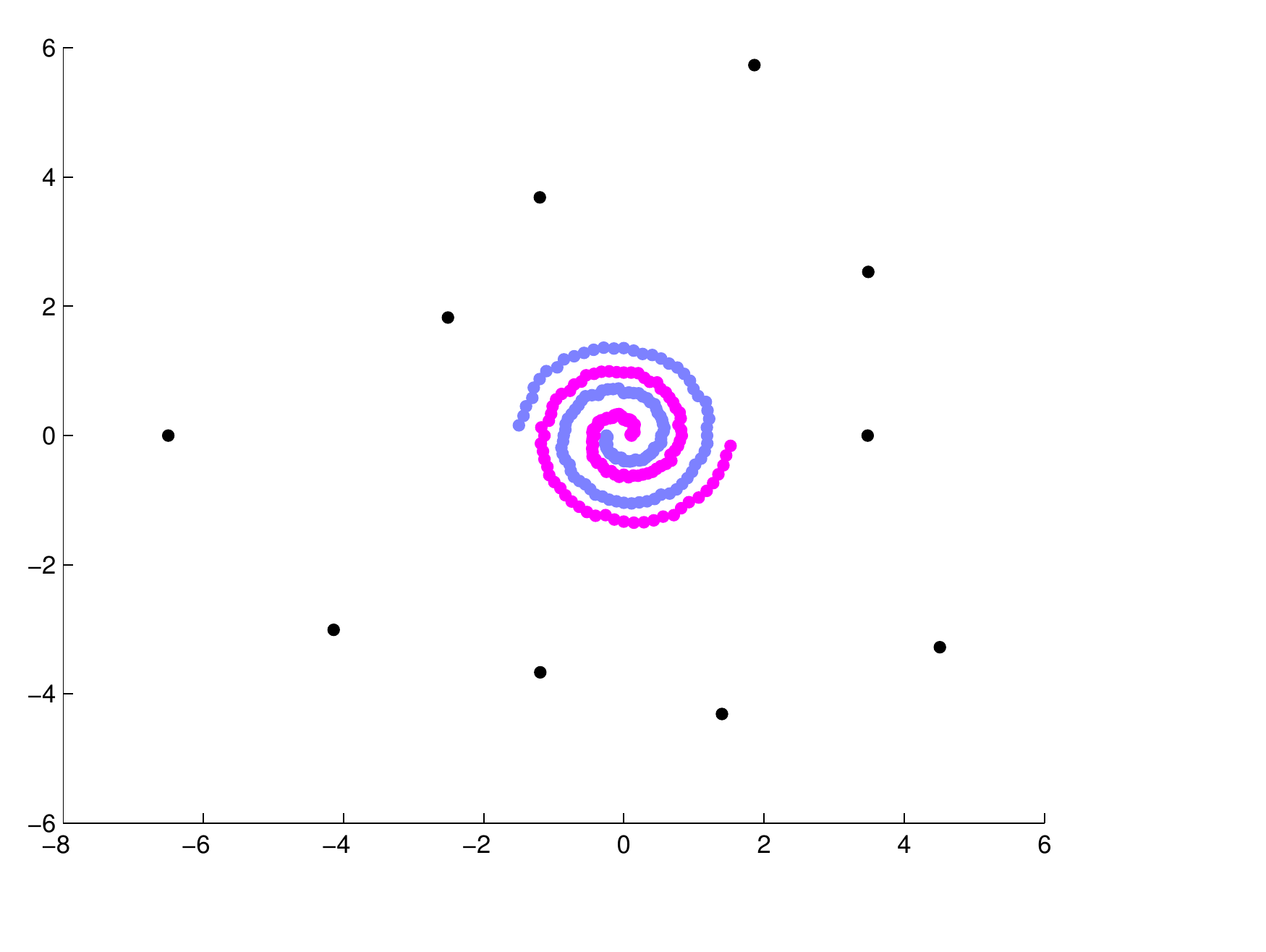}\\ 
       
        \end{tabular}
        
\end{figure}

Since it is a hard problem to extract the whole outlier profile of a graph, we use our clustering algorithm
to find an approximate minimizer for each $\alpha$ and we also apply a binary search on the spectrum of $\alpha$
to extract $I_*$ and $\alpha_*$  up to a predefined precision. Such an approximation algorithm is described in 
Algorithm~\ref{alg:spectheu}. It is important to note that based on what follows we can deduce that the performance 
of our algorithm is acceptable despite approximation, however the effect of approximation on the final result 
are undeniable (e.g. see Figure~\ref{fig:hardoutlier2}). Also, it should be noted that making the precision factors 
finer (e.g. smaller steps in the binary search) will give rise to  a longer runtime which is undesirable.

\begin{figure}[ht]
        \caption{Performance of Algorithm~\ref{alg:spectheu} (black spots are extracted outlier data points and     $\sigma_s$ = 0.1 for all problems).}
        \centering
        \vspace{10pt}
        \begin{tabular}{|c|c|}
        \hline
        \includegraphics[scale=.2]{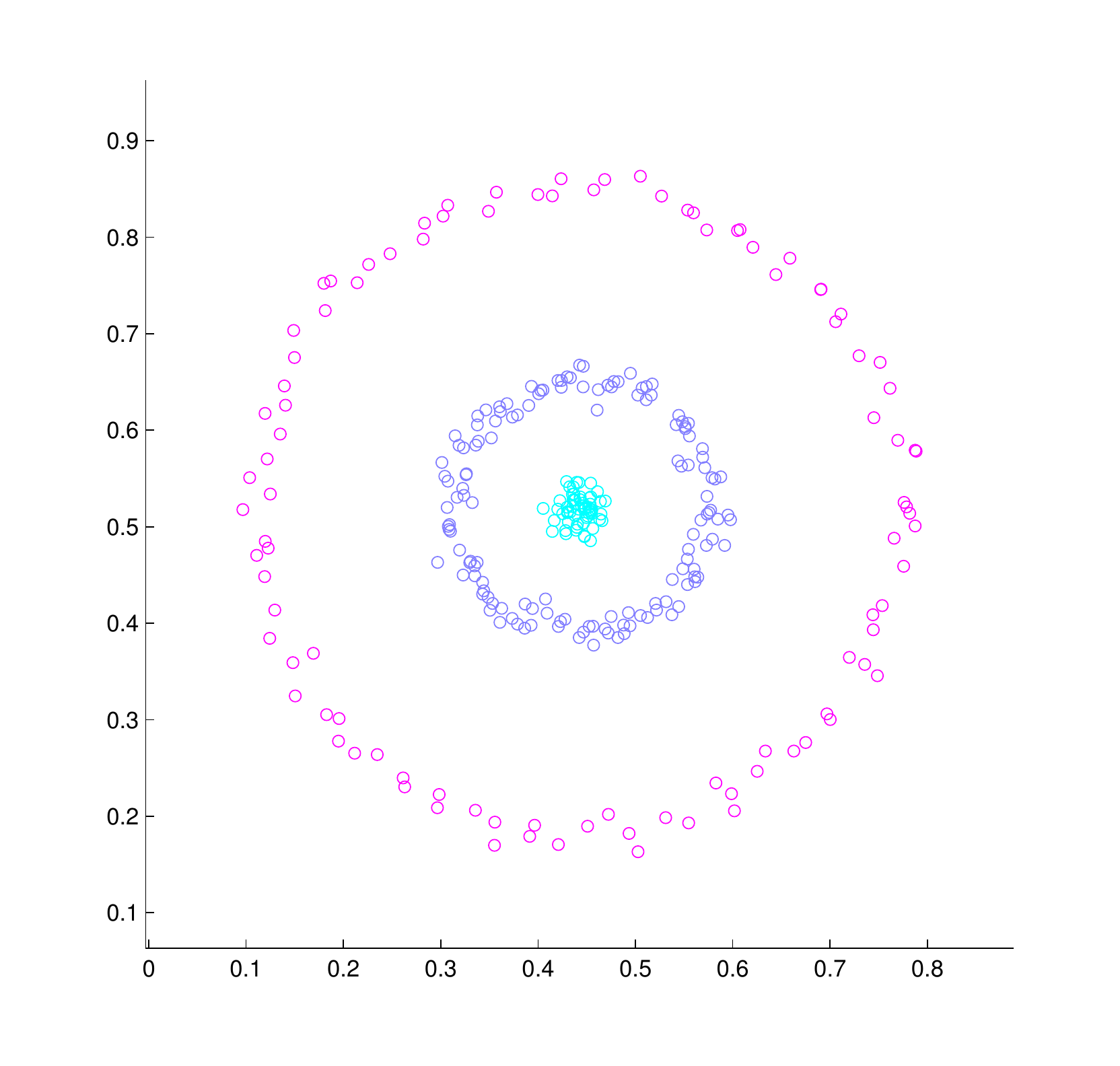}&\includegraphics[scale=.2]{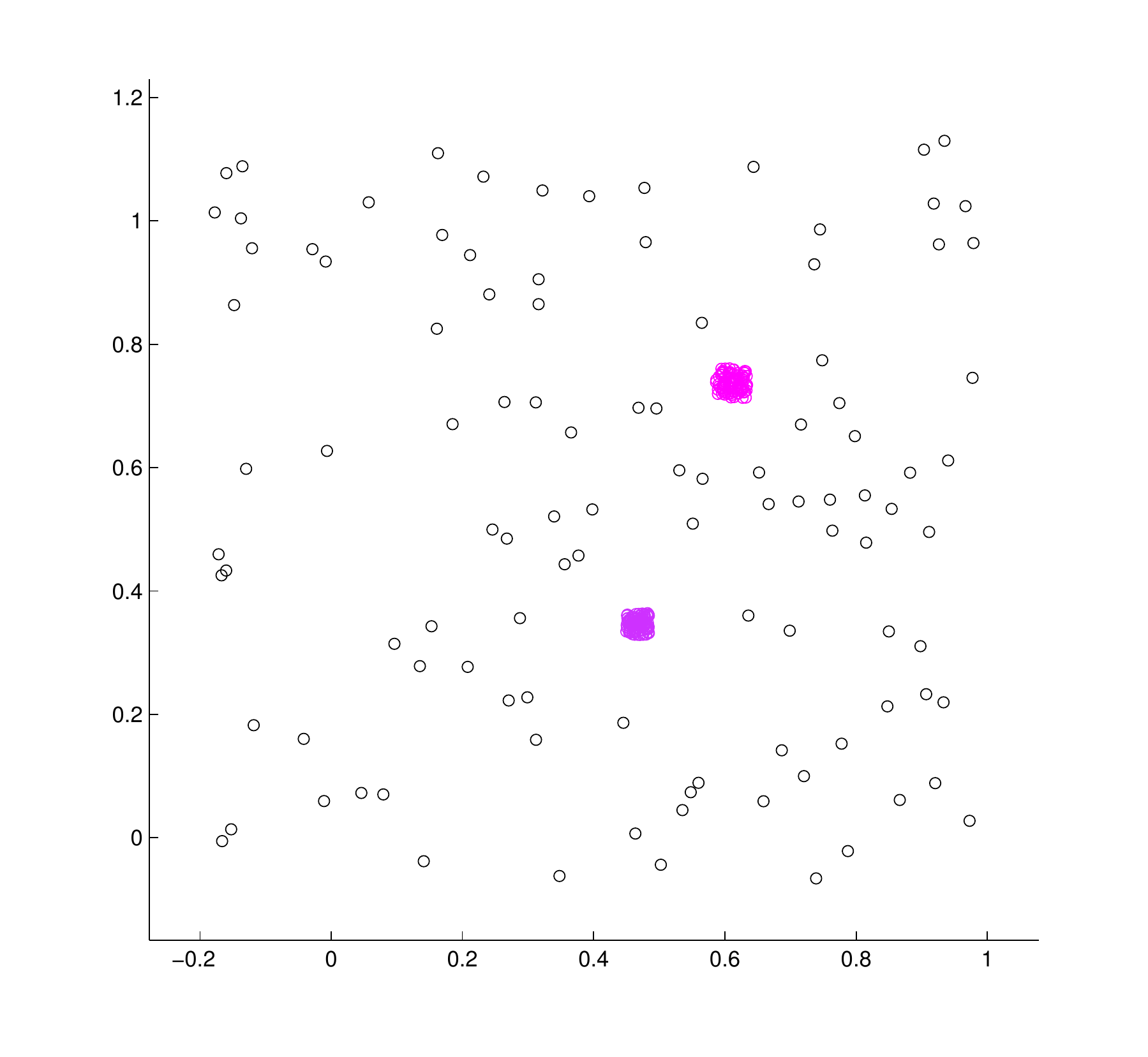}\\ \hline
        \newline
        \includegraphics[scale=.2]{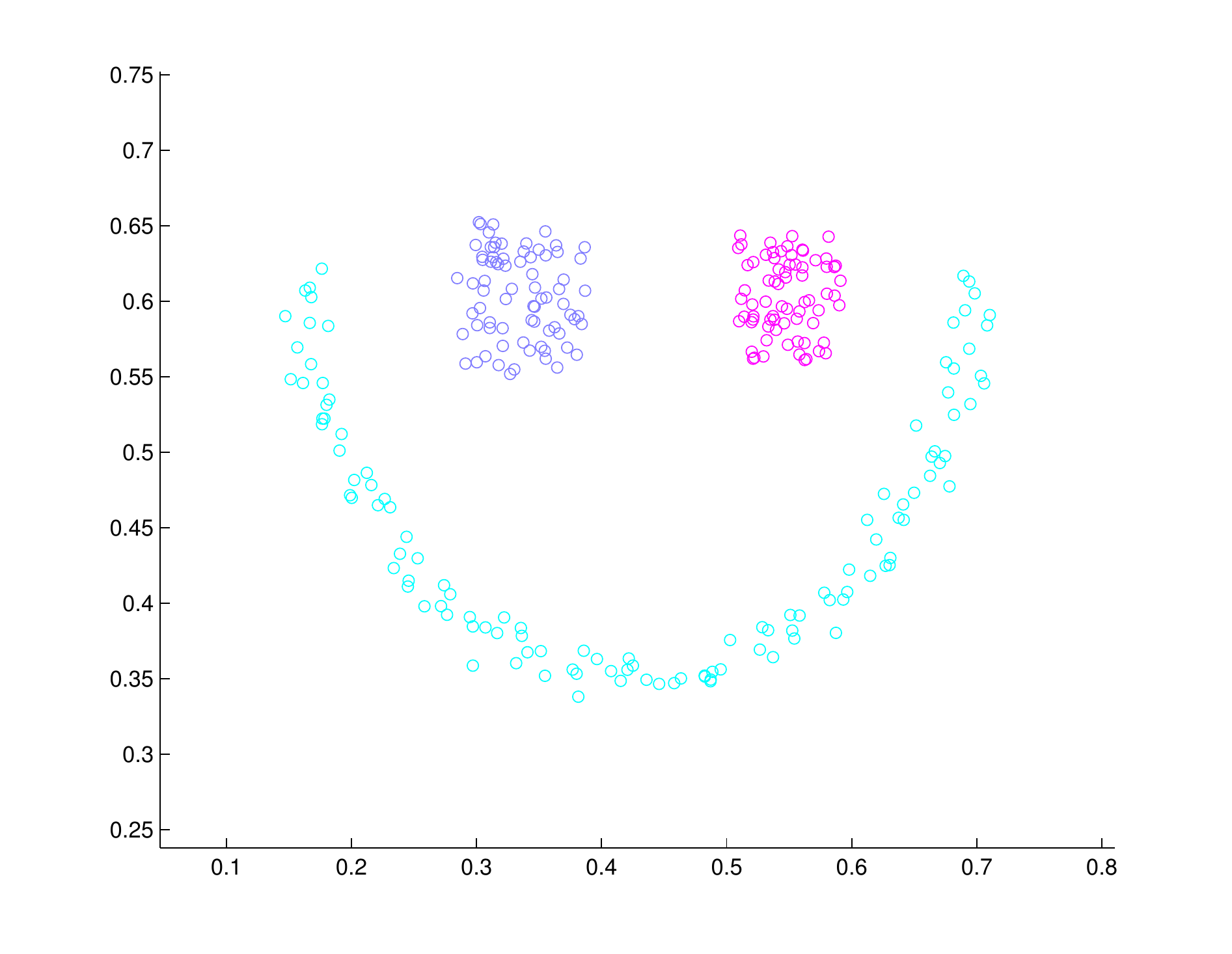}&\includegraphics[scale=.2]{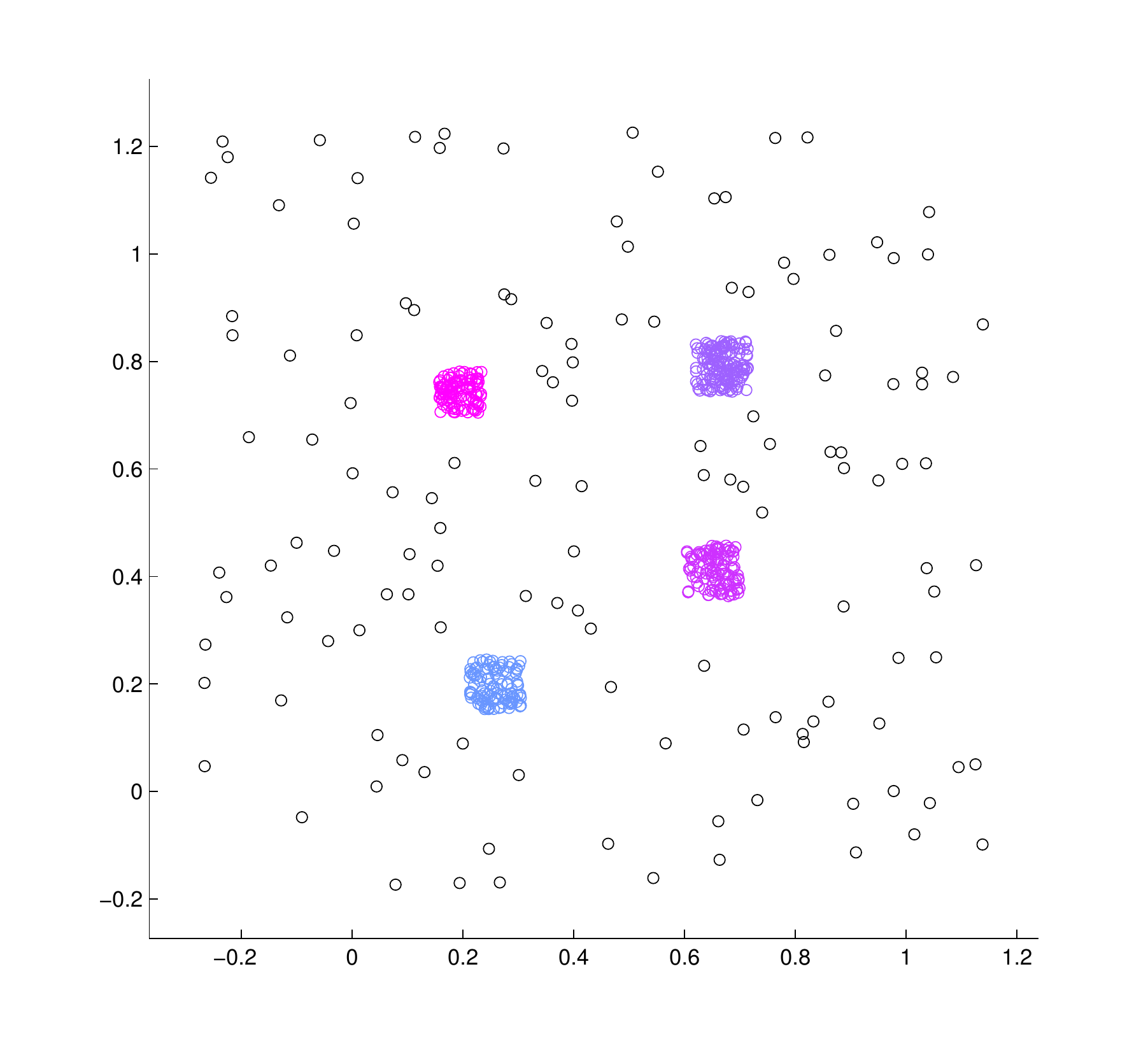}\\ \hline
        \newline
        \includegraphics[scale=.2]{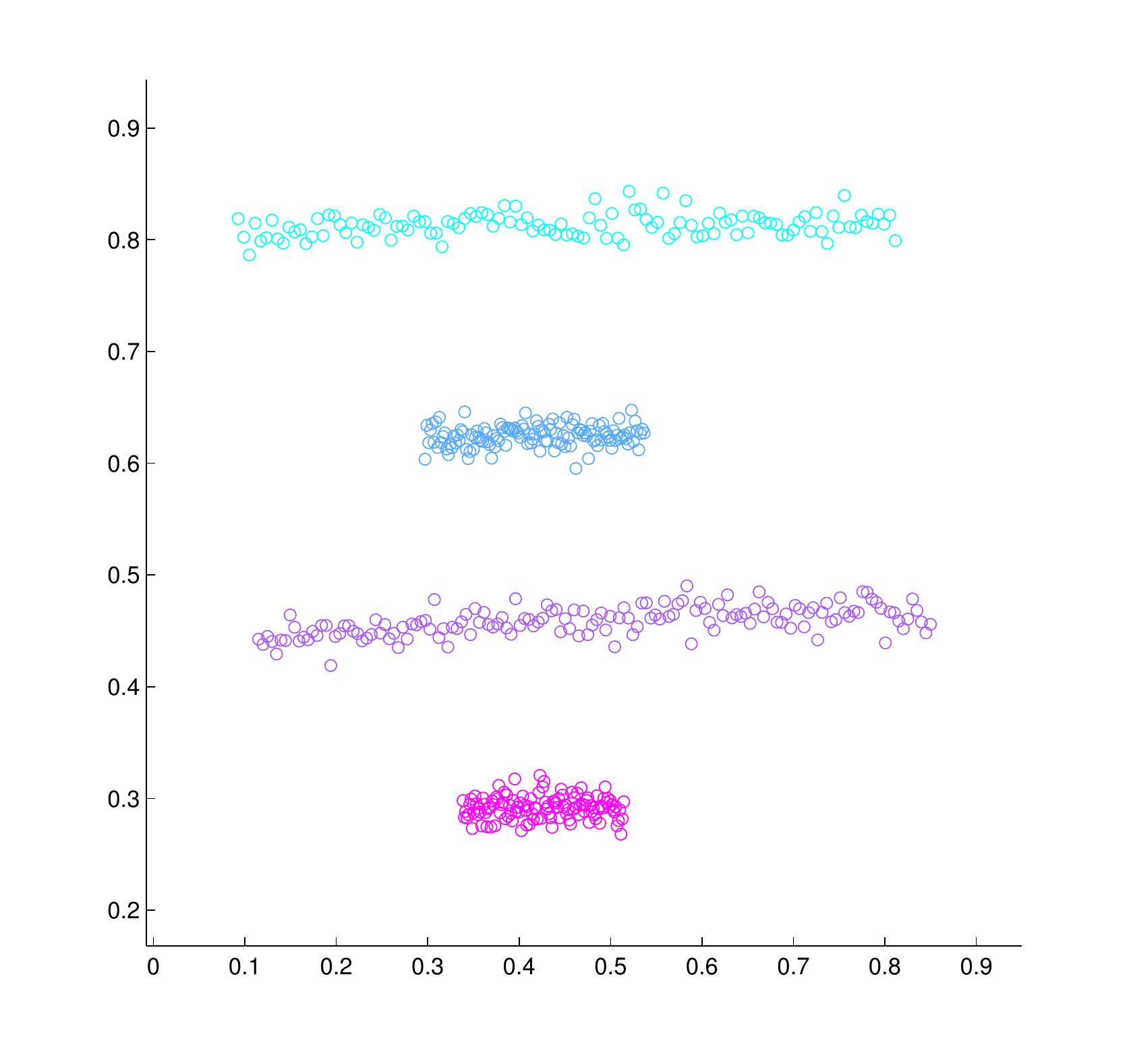}&\includegraphics[scale=.2]{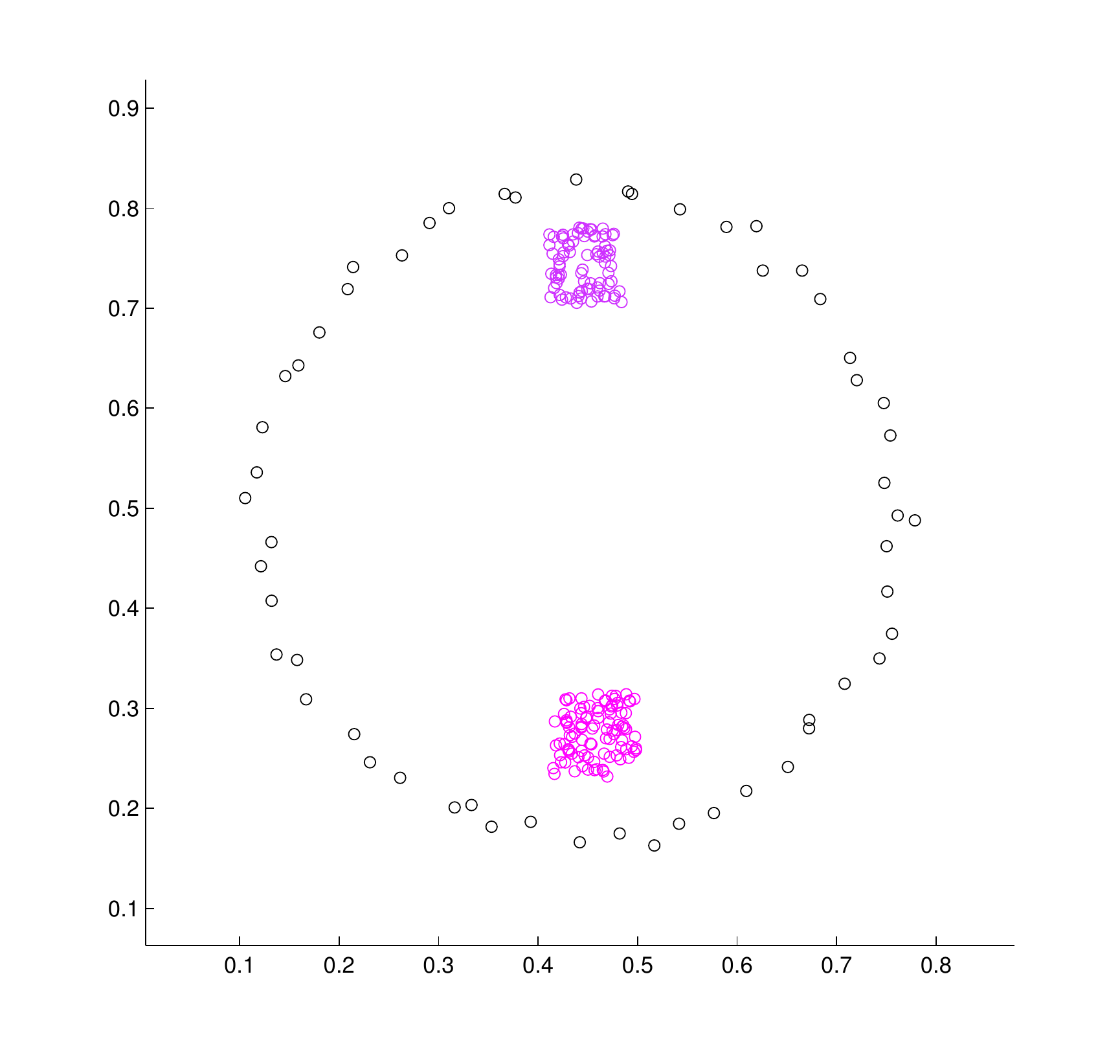}\\ \hline
        
        \end{tabular}
        \label{fig:caltechspectrum}
\end{figure}

In order to evaluate the performance of the proposed algorithm we consider a couple of artificial problems.
At first we applied the algorithm to the artificial clustering problems in Figure~\ref{fig:caltechspectrum}.
As it is clear from the results, the algorithm has been able to correctly extract clusters as well as the outlier set.

\begin{figure}[ht]
        \caption{\label{fig:hardoutlier}A problem with global affinity $\sigma = 0.09$ and $\sigma_s = 0.5$. 
        First row contains the outcome of WJHZQ \cite{WA08} Algorithm. Second row contains the outcome of
        our algorithm with $\alpha_* = 0.25$. }
        \centering
        \begin{tabular}{c}
        \includegraphics[scale=.5]{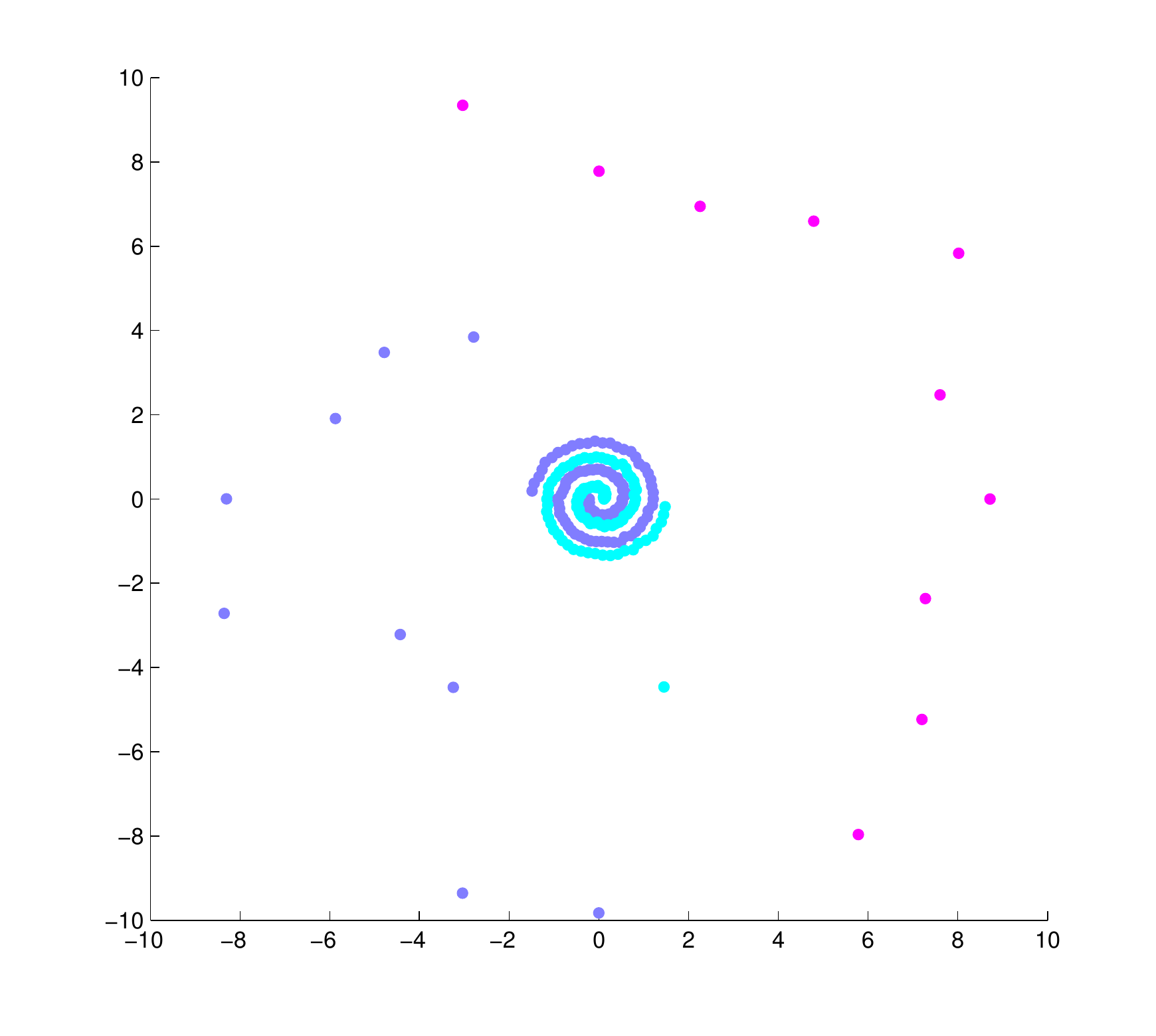}\\
        \includegraphics[scale=.5]{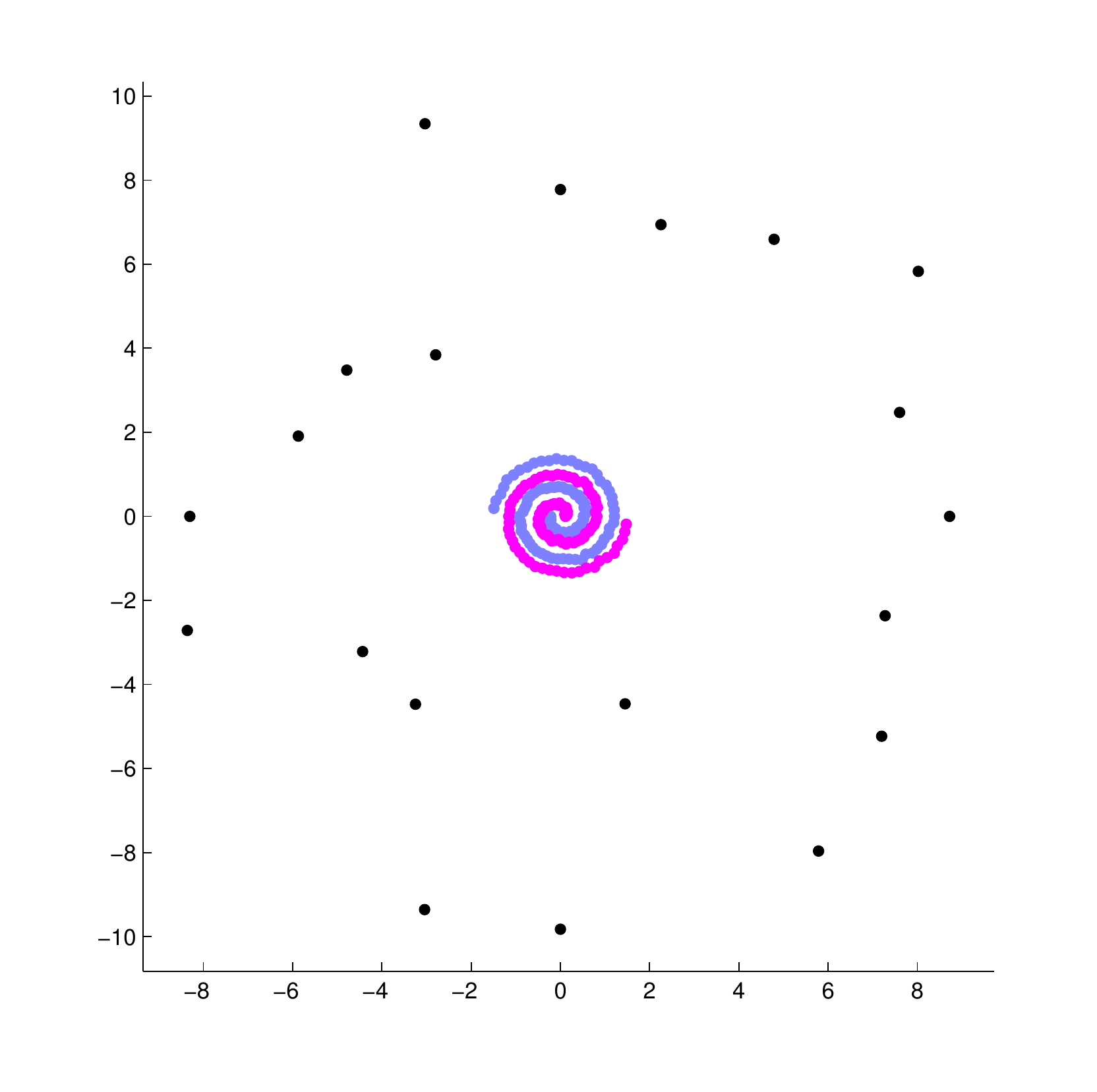}\\ 
       
        \end{tabular}
        
\end{figure}

In our final experimental evaluation we focus on the performance of the algorithm of Wang {\it et. al.} \cite{WA08} since it has 
the best performance according to our analysis in Section~\ref{sec:tests} and also since using a proper local scaling 
parameter it is capable of extracting the outlier set as a cluster.
In this direction we first consider a hard artificial clustering problem with outliers as depicted in Figure~\ref{fig:hardoutlier}. As it is clear from the results our algorithm has been able to successfully extract the outliers.

On the other hand, in order to test the impact of the local scaling on the performance we also consider a hard 
artificial clustering problem as depicted in Figure~\ref{fig:hardoutlier2} and as it is clear there are scaling parameters for which we can successfully extract the outlier set. In this regard it should be noted that in the local scaling setting for each individual case one can always find scaling parameters that works well, however, in our approach one may use the algorithm with as large a local scaling parameter as possible (considering ones runtime limits) without being concerned about the effect of this parameter on the performance of outlier extraction, while in 
the rest of the algorithms it is essentially this parameter which somehow tunes the algorithm to extract the outlier set as a cluster which is not consistent when one is dealing with a variety of different  data sets.

%% file: appendix.tex
Consider the following problem which is well-known to be $NP$-complete in the strong sense \cite{GJ79}. 

\begin{prob}{3-PARTITION}
{A positive integer $B$ and $3m$ positive integers $w_1,\ldots,w_{3m}$, such that $B/4< w_i< B/2$, for each $1\leq i\leq 3m$ and $\sum_{i=1}^{3m} w_i=mB$.}{Is there an $m$-partition $\{S_i\}^m_1 \in {\mathcal P}_m([3m])$ such that, for each $1\leq j\leq m$, $\sum_{i\in S_j} x_i=B$?}
\end{prob}

We provide a reduction from the 3-PARTITION problem. Assume that the integers $w_1,\ldots,w_{3m}$ together with the integer $B$  is an instance of the problem 3-PARTITION. Let $t$ be a fixed positive integer and construct the weighted tree $T=(V,E,\w,\ph)$ as follows.
\begin{align*}
V&:=\{x,x_i,y_j,z_l\ |\ 1\leq i\leq 3m, 1\leq j\leq m, 1\leq l\leq t\},\\
E&:=\{xx_i, xy_j, xz_l\ |\ 1\leq i\leq 3m, 1\leq j\leq m, 1\leq l\leq t\}.
\end{align*}
Vertex weights are defined as follows.
\begin{align*}
\w(x):=1,&\ \w(x_i):=w_i+B+1,\ \forall\ 1\leq i\leq 3m,\\
&\ \w(y_j):=1,\ \forall\  1\leq j\leq m,\\
&\ \w(z_l):=B+1,\ \forall\   1\leq l\leq t.
\end{align*}
All edge weights is set to be equal to $1$. Let $k=m+t$ and $N=1$ to get an instance of the problem MINIMUM RESIDUE NUMBER. We assume that $t$ is sufficiently large (e.g. $t>7m$). We are going to show that $\iso_k(T)={1}/{(B+1)}$. First note that for the subpartition $\cA:=\{\{x_1\},\ldots,\{x_{3m}\},\{z_1\},\ldots,\{z_t\}\}$, we have ${\rm cost}(\cA)={1}/{(B+1)}$ and thus $\iso_k(T)\leq 1/(B+1)$. On the other hand, let $\cB:=\{B_1,\ldots,B_k\}$ be a minimizing subpartition achieving $\iso_k(T)$. Then there is some $B_i$ which is completely included in the set $\{z_1,\ldots,z_t\}$. Thus ${\rm cost}(\cB)\geq 1/(B+1)$. This shows that $\iso_k(T)=1/(B+1)$.

Now assume that the answer to 3-PARTITION is positive and let $\{S_1,\ldots,S_m\}$ be a partition of $[3m]$, where the sum of the elements of each $S_i$ is equal to $m$. Therefore each $S_i$ has exactly 3 elements. Now define the $k-$subpartition $\cA:=\{A_1,\ldots,A_k\}$ as follows,
\begin{align*}
A_j&:=\{y_j,x_i\ |\ i\in S_j\}, \ \forall\ 1\leq j\leq m,\\
 A_j &:=\{z_{j-m}\}, \ \forall\ m+1\leq j\leq t+m.
\end{align*}
We have $\cA$ is a minimizing subpartition achieving $\iso_k(T)=1/(B+1)$ whose residue number is $1$.

Now, conversely, assume that there exists  a minimizing subpartition $\cB$ achieving $\iso_k(T)=1/(B+1)$ whose residue number is at most $1$.
Then the vertex $x$ is a residue element for $\cB$. Because for each $i$, $|B_i|\leq 3m+2$ and if $x\in B_1$, then ${\rm cost}(\cB)\geq t/ 2(3m+1)(B+1)>1/(B+1)$. 
Hence $\cB$ is a partition of the set $V\setminus\{x\}$. Since ${\rm cost}(\cB)=1/(B+1)$, for all $1\leq i\leq k$, the average of the weights of vertices in $B_i$ is at least $B+1$. Now delete all vertices $z_1,\ldots,z_t$ from $B_i$'s to obtain $m$ nonempty subsets $B'_1,\ldots,B'_m$ with the average weights of at least $B+1$. Since the average weight of all vertices $\{x_i,y_j\ | \ 1\leq i\leq 3m, 1\leq j\leq m\}$ is $B+1$, the average weight of each $B'_i$ is exactly $B+1$. Now, since $w_i$'s are positive, each $B'_i$ contains exactly one of the vertices $y_1,\ldots,y_m$. Hence for each $1\leq i\leq m$, $\sum_{x_p\in B'_i}w_p=B$. This finishes the reduction.